\documentclass[11pt]{article}
\usepackage{natbib} 

\usepackage[preprint]{acl}
\usepackage{booktabs}
\usepackage{siunitx}
\usepackage{float}
\usepackage{booktabs}
\usepackage{tabularx}
\usepackage{multirow}
\usepackage{threeparttable}
\usepackage{xcolor}
\usepackage{soul}
\usepackage{tikz}
\usepackage{placeins} 
\usetikzlibrary{arrows.meta, positioning}
\usepackage{amsmath}
\usepackage{amssymb}
\usepackage{subcaption} 

\usepackage{caption} 

\usetikzlibrary{shapes.geometric, arrows.meta, positioning, fit, backgrounds, calc, shadows, decorations.pathreplacing}

\tikzset{
    contribution box/.style={
        rectangle, rounded corners=3mm, 
        draw=#1, line width=2pt, 
        fill=#1!8,
        text width=10cm, 
        minimum height=11cm,
        align=left,
        drop shadow={opacity=0.25, shadow xshift=2pt, shadow yshift=-2pt}
    },
    title box/.style={
        rectangle, rounded corners=2mm,
        draw=#1, line width=2pt, fill=#1,
        text=white, font=\sffamily\bfseries\Large,
        inner sep=6pt,
        minimum width=10cm
    },
    number badge/.style={
        circle, draw=#1, line width=2.5pt, fill=white,
        text=#1, font=\sffamily\bfseries\Huge,
        minimum size=1.2cm,
        drop shadow={opacity=0.3}
    },
    arrow/.style={
        -Stealth, line width=2pt, draw=darkgray!60
    }
}

\definecolor{boxbg}{RGB}{250, 250, 250}
\definecolor{highlight}{RGB}{255, 243, 205}
\definecolor{openended}{RGB}{227, 242, 253}
\definecolor{openendedtext}{RGB}{21, 101, 192}
\definecolor{closedended}{RGB}{243, 229, 245}
\definecolor{closedendedtext}{RGB}{106, 27, 154}
\definecolor{primaryblue}{RGB}{102, 126, 234}
\definecolor{primarypurple}{RGB}{118, 75, 162}
\definecolor{truegreen}{RGB}{40, 167, 69}
\definecolor{falsered}{RGB}{220, 53, 69}
\definecolor{ignoregray}{RGB}{108, 117, 125}
\definecolor{partialorange}{RGB}{253, 126, 20}
\definecolor{strictpurple}{RGB}{111, 66, 193}
\definecolor{lightgray}{RGB}{248, 249, 250}
\definecolor{candidateyellow}{RGB}{255, 193, 7} 
\definecolor{questioncyan}{RGB}{23, 162, 184} 
\definecolor{gtgreen}{RGB}{40, 167, 69}
\definecolor{contribution1}{RGB}{70,130,180}    
\definecolor{contribution2}{RGB}{220,20,60}     
\definecolor{contribution3}{RGB}{46,139,87}     
\definecolor{lightblue}{RGB}{173,216,230}
\definecolor{lightred}{RGB}{255,182,193}
\definecolor{lightgreen}{RGB}{144,238,144}
\definecolor{dgray}{RGB}{64,64,64}
\sethlcolor{highlight}
\usepackage{pifont}
\newcommand{\cmark}{\textcolor{truegreen!60!black}{\ding{51}}}
\newcommand{\xmark}{\textcolor{falsered!80!black}{\ding{55}}}
\usepackage{array}
\usepackage{tabularx}

\newcolumntype{Y}{>{\raggedright\arraybackslash}X}

\usepackage{hyperref}
\usepackage{url}

\usepackage{times}
\usepackage{latexsym}

\usepackage[T1]{fontenc}

\usepackage[utf8]{inputenc}

\usepackage{microtype}

\usepackage{inconsolata}

\usepackage{graphicx}
\usepackage[most]{tcolorbox} 
\usepackage{enumitem} 
\usepackage{caption}
\usepackage{setspace}
\usepackage{booktabs}

\title{PHANTOM RECALL: When Familiar Puzzles Fool Smart Models}

\author{
Souradeep Mukhopadhyay$^{1*}$ \quad Rishabh Baral$^{1*}$ \quad Nimeesh Mahajan$^{1*}$ \quad Samhitha Harish$^{1*}$\\
\textbf{Aswin RRV}$^{1}$ \quad \textbf{Mihir Parmar}$^{1}$ \quad \textbf{Mutsumi Nakamura}$^{1}$ \quad \textbf{Chitta Baral}$^{1}$\\
\\
$^1$School of Computing and Augmented Intelligence, Arizona State University\\
\texttt{\{smukho11, rbaral1, mutsumi, chitta\}@asu.edu}
}

\begin{document}
\maketitle
\begin{abstract}
Large language models (LLMs) such as GPT, Gemini, and Claude often appear adept at solving classic logic puzzles—\textit{but how much genuine reasoning underlies their answers?} Recent evidence suggests that these models frequently rely on memorized templates rather than reasoning from first principles. When puzzles are slightly modified, their performance collapses, revealing a striking fragility. In particular, we asked: \textit{Have LLMs addressed these issues? To what extent? How about perturbations to other puzzles?Is there a general way of reformulating the prompt so that the models do better?} To examine these things systematically, we introduce \textbf{PHANTOM RECALL}, a benchmark comprising 25 well-known logic puzzles and 149 carefully designed perturbations that preserve reasoning structure but alter superficial details and solutions. We evaluate eleven leading LLMs and identify a recurring failure mode--phantom recall--where models confidently reproduce memorized solutions or spurious rationales that no longer fit the altered scenario. To probe and mitigate this issue, we contribute three tools: (i) an automated logical-equivalence judge to detect reasoning mismatches, (ii) a taxonomy of fine-grained reasoning error categories, and (iii) a prompting-based mitigation framework guided by these categories. Despite near-perfect accuracy on unmodified puzzles, models significantly underperform humans on perturbed ones, exhibiting both phantom recall and over-elaboration. Our findings reveal a crucial limitation: LLMs often fail to re-reason when contextual cues shift--highlighting the gap between linguistic fluency and logical understanding.

\def\thefootnote{$*$}\footnotetext{\footnotesize Equal Contributions}\def\thefootnote{\english{footnote}}


\end{abstract}

\section{Introduction}
\begin{figure}[!h]
    \centering
    \includegraphics[width=1\linewidth]{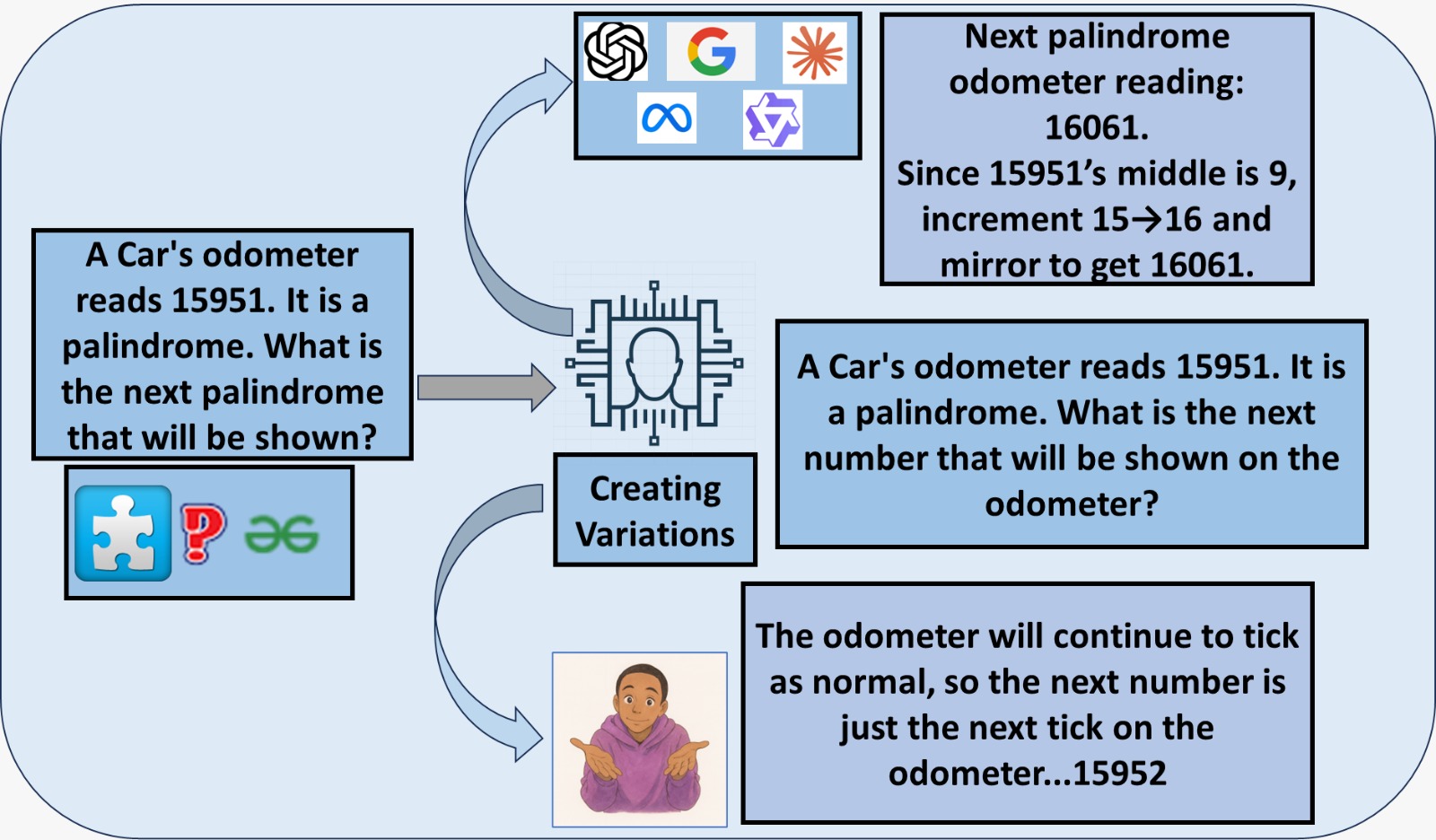}
    \caption{LLMs stumbled on this simple trivial puzzle variant--but a human can solve it instantly.}
    \label{rr}
\end{figure}

It began with a puzzle--a riddle so simple that any schoolchild could solve it. Yet when posed to a large language model (LLM), the model answered with confidence--and was completely wrong. This amusing failure points to something deeper: despite their fluent prose and encyclopedic recall, LLMs often falter on tasks that require precise reasoning and faithful interpretation of instructions. \citet{marcus2024alphaproof} catalogs numerous such failures, including the viral "strawberry problem" where models miscounted letters in a word—a task trivial for humans but revealing fundamental gaps in character-level reasoning. Statistician Colin Fraser has systematically documented this brittleness by "torturing" models with slight variations on familiar problems, demonstrating that minor perturbations often cause dramatic performance collapse as he shows in his twitter demonstrations on GPT Fails \cite{fraser2024gpt}.

Over the past year, social media has amplified this phenomenon. Users frequently share screenshots of models stumbling on basic logic questions, misreading constraints, or hallucinating details never asked for. At first glance, these seem like trivial mistakes, but systematic studies confirm the pattern: even slight changes in wording, the addition of a constraint, or a twist on a familiar question can sharply degrade model accuracy \cite{srivastava2023beyond, lightman2023let}. These failures hint that models may be \textit{pattern matchers rather than genuine reasoners}, leaning on statistical echoes of training data rather than recomputing solutions from first principles.

This issue is further complicated by benchmark contamination. Analyses reveal that benchmark familiarity can mask these weaknesses: models may reproduce memorized solutions to problems that appeared in training data, inflating apparent performance \cite{golchin2023benchmarks}. Alignment effects such as \textit{sycophancy} exacerbate the problem, with models optimizing for agreement and style over accuracy, often producing confident yet logically incoherent answers \cite{perez2022discovering}.

In this work, we turn this anecdotal observation into a rigorous evaluation paradigm. Inspired by Fraser's approach of systematic perturbation \cite{fraser2024gpt}, we compile a diverse set of riddles and logic puzzles, systematically perturb their constraints (basically changing the question to be as trivial as possible), and observe whether LLMs adjust their reasoning. We additionally construct “mirror image” variants designed to constrain the answer space, coaxing the model toward a correct solution with minimal wording changes. Our guiding question is simple: \textbf{when the puzzle changes, does the reasoning change too?} Our exploration of this main question yields a resounding answer: \textbf{More often than not, the model does not change its reasoning, leading to often humorous failures in logic on even the simplest of problems.}The overall problem scenario is illustrated in Figure \ref{rr}. Our paper makes three major contributions:
\begin{enumerate}[noitemsep]
    \item 
    We introduce a controlled suite of 25 classic puzzles with 149 carefully constructed variants that preserve the underlying reasoning template while changing the correct answer, revealing a recurring failure mode--\emph{phantom recall}--where LLMs reproduce stock solutions that don’t fit the instance. This isolates instance-level reasoning robustness under minimal perturbations.

    \item 
    Beyond final-answer accuracy, we provide (i) an automated conceptual-equivalence judge with answer verification auto-evaluator, (ii) a lightweight taxonomy of thinking-pattern configurations with fine-grained error labels (e.g., deductive failure, cascading error, improper elimination), and (iii) an LLM-aided step error classification autoevalutor
that mirrors human step-analysis to surface where and why chains break—enabling model-agnostic, low-effort diagnosis of reasoning behavior.

    \item 
    Across 11 LLMs (6 closed models and 5 open-source models), SOTA systems approach perfect accuracy on base puzzles but drop sharply on perturbed variants, with errors dominated by phantom recall and over-elaboration; “thinking/structured” prompting reliably improves outcomes yet does not close the robustness gap--highlighting concrete targets for future training-time interventions.
\end{enumerate}

\section{Related Work}
\subsection{Puzzle and its variation solving}
Puzzle-solving offers a window into LLMs’ logical abilities. \citet{Giadikiaroglou2024PuzzleSU} divide puzzles into (1) rule-based and (2) rule-less categories. Rule-less types include riddles \cite{Lin2021RiddleSenseRA}, MCQs \cite{Zhao2023SolvingAG}, programming puzzles \cite{Schuster2021ProgrammingP}, and commonsense reasoning puzzles \cite{Gu2023BeyondTO}; our focus is on rule-based puzzles. Prior efforts on rule-based puzzles span Sudoku \cite{Noever2021PuzzleSW}, Rubik’s Cube, the 8-puzzle, Game of 24 \cite{Yao2023TreeOT}, crosswords \cite{Yao2023TreeOT}, chess puzzles \cite{Feng2023ChessGPTBP}, card games \cite{Gupta2023AreCA}, BoardgameQA \cite{Kazemi2023BoardgameQAAD}, and Lateral Thinking Puzzles \cite{Huang2023LatEvalAI}. Similarly, \citet{Mitra2015LearningTA} proposed a grid-based puzzle dataset, and \cite{Dziri2023FaithAF} studied compositionality in LLMs using Grid Puzzle. Besides that, some minimal works have been done on compositional variation on puzzle. \citet{Williams2024EasyPT} showed that LLMs can easily get wrong on the puzzle variation. But reasoning analysis of why LLMs can easily get wrong, has not been done yet. Getting motivated by that, we want to explore the detailed reasoning chain, where LLMs are getting wrong with a proposed puzzle variation dataset.

\subsection{Step-by-Step analysis of reasoning chains}
Reference-free evaluation, which is independent of gold reasoning chains, has been the focus of earlier studies \cite{Dalvi2021ExplainingAW, Saparov2022LanguageMA, Han2022FOLIONL}. ROSCOE \cite{Golovneva2023ROSCOEAS} provides a broad metric suite to evaluate reasoning quality across correctness, informativeness, consistency, and coherence. ReCEval \cite{Prasad2023ReCEvalER} narrows the focus to incorrect-answer detection by analyzing the correctness and informativeness of intermediate steps. LLM Reasoners \cite{Hao2024LLMRN} offers a framework for fine-grained, step-by-step assessment of large language models’ reasoning. \citet{Ling2023DeductiveVO} introduces Natural Program, a natural-language deductive format that breaks verification into sequential substeps. \citet{Tyen2023LLMsCF} examines LLM limitations in spotting reasoning errors while showing strong performance at fixing them when error locations are provided. \citet{Li2022MakingLM} presents DIVERSE, a three-stage method to identify and correct errors throughout a reasoning chain. Although prior work proposes valuable evaluation frameworks \cite{Patel2024MultiLogiEvalTE,Parmar2024LogicBenchTS,Tyagi2024StepbyStepRT,Mishra2025InvestigatingTS}, they are not tailored to puzzle variation. In contrast, our approach employs simple yet effective soundness and correctness metrics to yield detailed insights into step-wise standard puzzle variation reasoning errors and to enable scalable, low-effort evaluation via an LLM-based framework.

\section{Dataset}
\subsection{Dataset Construction}
\begin{figure}
    \centering
    \includegraphics[width=1\linewidth]{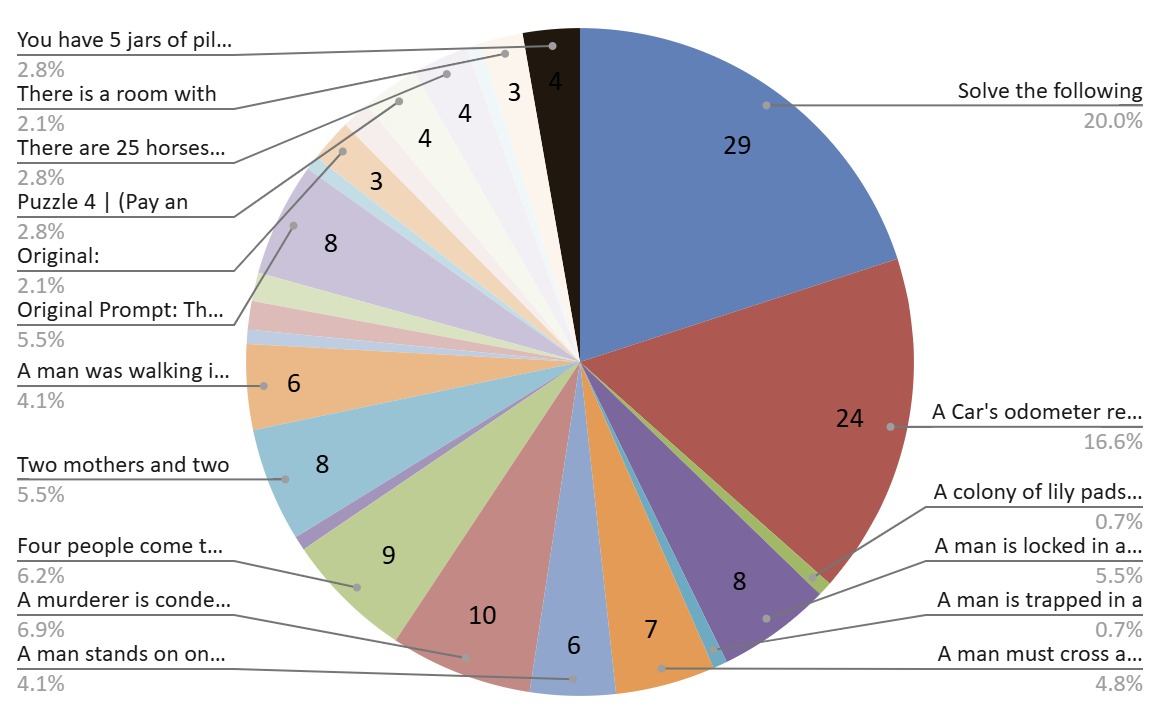}
    \caption{Distribution of standard puzzle and variations.}
    \label{fig:dataset}
\end{figure}

We began with a base set of 25 classic riddles and logic problems, sourced from a combination of Wikipedia entries on famous puzzles, online puzzle forums, and brainteaser collections. To diversify the set, we also generated riddles with ChatGPT based on human-authored examples, then manually reviewed and edited them before inclusion. For each base puzzle, we manually created \textbf{perturbed variants} by altering constraints, numerical values, or other problem details. The key principle was to preserve the \textit{reasoning structure} while forcing a different final answer, ensuring that models could not rely on memorization. These perturbations allow us to test whether models genuinely recompute solutions rather than simply recalling previously seen ones. Pie chart of Figure \ref{fig:dataset} shows the number of standard puzzle and variations. 
\subsection{Mirror-Image Dataset}

To further probe reasoning, we developed a \textbf{mirror-image subset} of the dataset. Here, 43 of the more open-ended puzzle variations were rewritten to constrain the possible answer space. For example, an open-ended riddle asking “What could the object be?” might be reformulated into a multiple-choice format with three plausible candidates. This allows us to measure whether additional structure or hints can recover correct reasoning when the free-form version fails. It is described in App. \ref{app:question_transformation} in a great details with several examples.

\subsection{Validation and Quality Control}

Every riddle and its variants underwent a multi-step validation process:

\begin{enumerate}[noitemsep]
    \item \textbf{Instruction-Following Check:} We verified that models interpreted the question correctly rather than answering an unintended reformulation.
    \item \textbf{Reasoning Efficiency:} We flagged cases where the model’s solution path was overly convoluted, suggesting inefficiency or spurious reasoning.
    \item \textbf{Ambiguity Resolution:} If a model produced an unintended but logically valid answer, we revised the puzzle wording to eliminate multiple interpretations.
    \item \textbf{Consistency Assessment:} We documented cases where models triggered self-verification (e.g., “Let me double-check”), as these often revealed reasoning uncertainty.
\end{enumerate}

\subsection{Answer Verification Auto-Evaluator}
\begin{figure}
    \centering
    \includegraphics[width=1\linewidth]{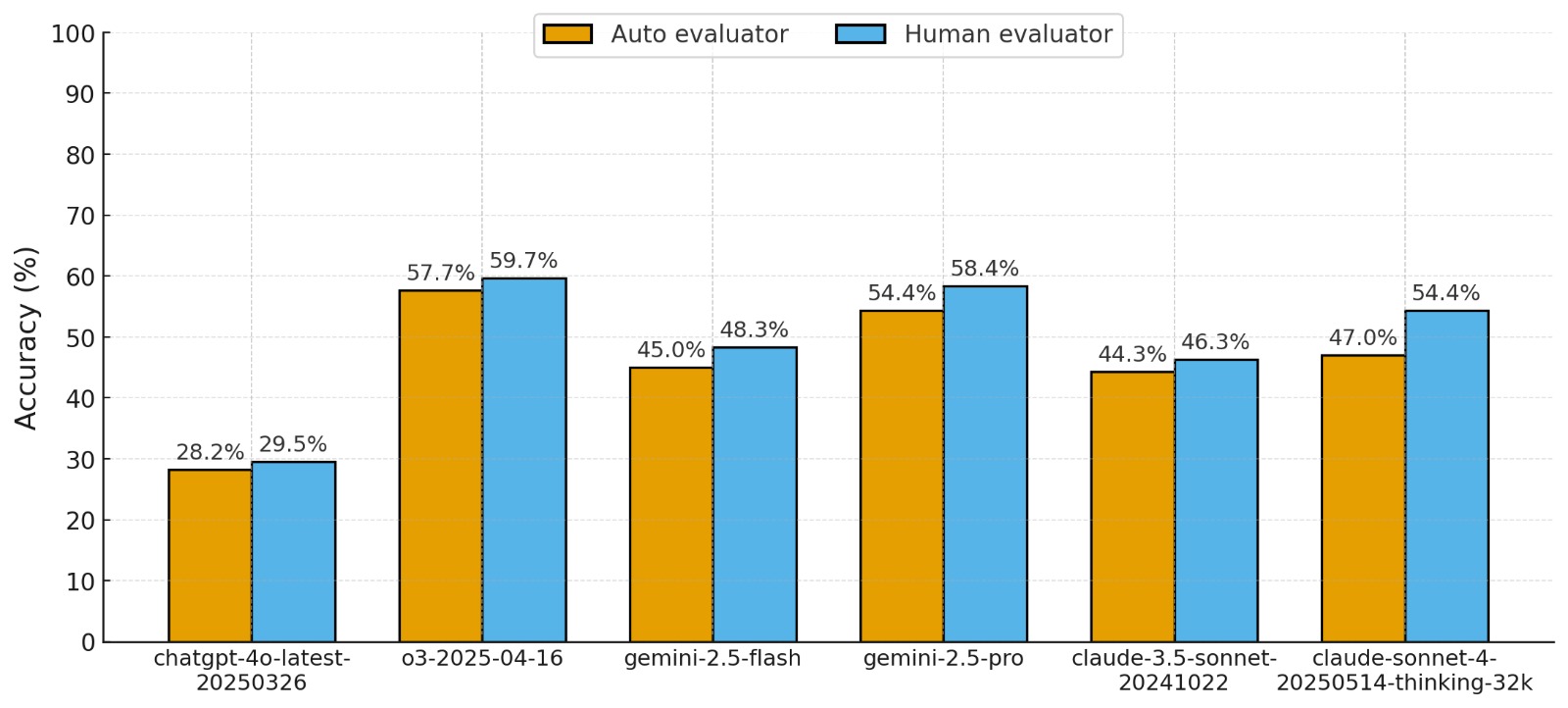}
    \caption{Grouped bar chart: Auto vs Human accuracy on closed source models}
    \label{fig:tt}
\end{figure}

We developed an LLM-based auto-evaluator to verify the final answer for each data point. Each predicted answer is compared with the gold answer. If the answer matches, the instance is assigned a label of \textit{TRUE}, and \textit{FALSE} otherwise. The human verification result is described in  App. \ref{app:human_eval_metrics}. Answer verification auto-evaluator system prompt is described in App. \ref{app:AAA}. The accuracy comparison on closed source model is shown in Figure \ref{fig:tt}. It states that our LLM based Answer verification auto-evaluator gives $\sim$96\% similarity with human annotations.

\section{Evaluation of Reasoning Chains}

\subsection{Manual Evaluation of Reasoning Chains}
To move beyond automated metrics and gain a deeper understanding of the failure modes in modern language model reasoning, we performed a rigorous qualitative analysis of their generated outputs. Our analysis corpus consisted of 50 reasoning chains generated for variation puzzles. These were sampled from five distinct open-source models: (LLaMa\cite{Touvron2023LLaMAOA}, InternLM\cite{Cai2024InternLM2TR}, Qwen\cite{Bai2023QwenTR}, Phi\cite{Abdin2024Phi3TR}, and Mistral\cite{Jiang2023Mistral7}), with 10 chains evaluated per model to ensure a comparative perspective. The lengths of the chains generated by the different Open-Source LLMs analyzed are discussed in the table in App. \ref{app:step_counts}. 
Our manual evaluation protocol involved two stages:

\begin{enumerate}[noitemsep]
    \item \textbf{Logical Step Segmentation:} First, each complete reasoning chain was deconstructed into its constituent sentences. Each sentence was treated as an individual reasoning step, allowing for a granular assessment of the model's logical progression.
    \item \textbf{Premise-Conclusion Analysis:} Second, each segmented step was meticulously examined to isolate its core logical components: the premise(s) upon which the reasoning was based, and the resulting conclusion. We then assessed the validity of this inference, determining whether the conclusion logically followed from the premise and whether the premise itself was factually correct given the puzzle's constraints and the preceding steps.
\end{enumerate}

The primary objective of this methodology was to create a comparative profile of reasoning errors, enabling us to identify systemic weaknesses and differentiate the qualitative performance of each model. Additionally, to ensure that the error categorization was performed with the utmost attention to detail, the human annotators were given the annotation guidelines in App. \ref{app:annotation_guidelines} to adhere to. 
\subsection{Proposed Error Categories}
Based on our analysis, we developed a framework to systematically classify the errors within each reasoning step. This framework consists of five high-level categories based on the soundness of the premise and conclusion:
\begin{itemize}[noitemsep]
    \item \textbf{Logically Sound Step:} A step where a correct conclusion is validly derived from a correct premise.
    \item \textbf{Deductive Failure:} A step where a correct premise leads to an incorrect conclusion due to flawed reasoning.
    \item \textbf{Compounded Failure:} A step where an incorrect premise leads to an incorrect conclusion.
    \item \textbf{Fortuitous Correctness:} An interesting case where a flawed premise coincidentally results in a correct conclusion.
    \item \textbf{Declarative Statement:} A sentence that restates a clue or a previous finding without drawing a new conclusion.
\end{itemize}

To pinpoint the root causes of failures, we further analyzed the origins of flawed premises and the nature of incorrect conclusions.
\textbf{Sources of Flawed Premises:} We identified two sources for incorrect premises: (1) \textbf{Evidence-based}, where information is drawn directly from the puzzle's clues, and (2) \textbf{Internally-derived}, where the premise is inferred from previous steps. Errors associated with these sources include:

\begin{itemize}[noitemsep]
    \item \textbf{Evidence Misrepresentation:} The premise fabricates or distorts factual information from the clues.
    \item \textbf{Insufficient Evidence:} The premise is factually correct but inadequate to support the drawn conclusion.
    \item \textbf{Unsupported Assumption:} The model introduces a speculative premise not directly given by or derived from the clues.
    \item \textbf{Cascading Error:} An error from a previous conclusion is carried forward as a flawed premise.
    \item \textbf{Faulty Postulate:} An explicitly stated assumption is derived incorrectly from prior information.
\end{itemize}

\textbf{Types of Flawed Conclusions:} We classified incorrect conclusions into three primary types:

\begin{itemize}[noitemsep]
    \item \textbf{Consequential Error:} The conclusion is incorrect as a direct result of a flawed premise.
    \item \textbf{Improper Candidate Elimination:} The model fails to correctly discard all invalid options, a common failure in constraint-based puzzles.
    \item \textbf{General Logical Fallacy:} A catch-all for other invalid deductions where the reasoning itself is erroneous, independent of the premise's correctness.
\end{itemize}

For clarity, we summarize this error taxonomy in Table \ref{tab:error_taxonomy}.
\begin{table*}[t]
\centering
\scriptsize  
\setlength{\tabcolsep}{3pt}
\renewcommand{\arraystretch}{0.68}
\begin{tabular}{p{2.6cm}|p{3.1cm}|p{2.6cm}|p{5cm}}
\toprule
\textbf{Category} & \textbf{Source} & \textbf{Sub-Category} & \textbf{Description} \\
\midrule
\multirow{5}{2.6cm}{Deductive Failure, Fortuitous Correctness, Compounded Failure, Declaration\vspace{-5pt}}& \multirow{3}{3.1cm}{From premise (e.g., 'Considering first two points...')\vspace{-3pt}}& (1) Evidence Misrepresentation& Premise fabricates or distorts factual information.\\
\cmidrule{3-4}
& & (2) Insufficient Evidence& Premise correct but inadequate to support conclusion. \\
\cmidrule{3-4}
& & (3) Unsupported Assumption& Model introduces premise not given by clues. \\
\cmidrule{2-4}
& \multirow{2}{3.1cm}{Derived Conclusions using premises not given.\vspace{-6pt}}& (4) Cascading Error& Error from previous conclusion carried forward. \\
\cmidrule{3-4}
& & (5) Faulty Postulate& Assumption derived incorrectly from prior info. \\
\midrule
\multirow{3}{2.6cm}{Compounded Failure}& \multirow{3}{3.1cm}{Derived using premise (taken directly or derived)}& (a) General Logical Fallacy& Invalid deductions with erroneous reasoning.\\
\cmidrule{3-4}
& & (b) Consequential Error& Conclusion incorrect due to flawed premise.\\
\cmidrule{3-4}
& & (c) Improper Candidate Elimination& Fails to discard invalid options. \\
\bottomrule
\end{tabular}
\caption{\textbf{Error taxonomy for reasoning failures.}}
\label{tab:error_taxonomy}
\end{table*}
\subsection{Step Error Classification auto-evaluator}
While our manual analysis provides a deep, qualitative understanding of reasoning failures, its meticulous nature makes it challenging to scale across our entire dataset. To analyze the distribution of errors at a larger scale--a crucial step for understanding the systemic shortcomings of LLMs in creative reasoning--we developed an LLM-based automated evaluator.

We termed this system the \textbf{"Step Error Classification auto-evaluator"}. It utilizes the GPT-4o model to automate the error identification and categorization process for riddle-solving reasoning chains. The auto-evaluator is guided by a comprehensive prompt containing:

\begin{enumerate}[noitemsep]
    \item \textbf{System Instructions:} A detailed set of rules and sequential steps, mirroring our manual evaluation protocol, that instruct GPT-4o on how to deconstruct and analyze a reasoning chain. This also specifies the required JSON output format for parsing.
    \item \textbf{Domain Knowledge:} A complete definition of our error classification framework, including the five high-level categories and their associated sub-categories. To minimize ambiguity, we provide a clear preference order for classifying errors that might fit multiple definitions.
    \item \textbf{A Guiding Exemplar:} A one-shot example consisting of a riddle, its ground truth answer, a model-generated reasoning chain, and our corresponding manual evaluation, which serves as a template for the expected output.
\end{enumerate}

The full prompt structure for the Auto-evaluator is provided in App \ref{app:auto_evaluator}.

\begin{figure*}[t]
    \centering
    \includegraphics[width=\textwidth]{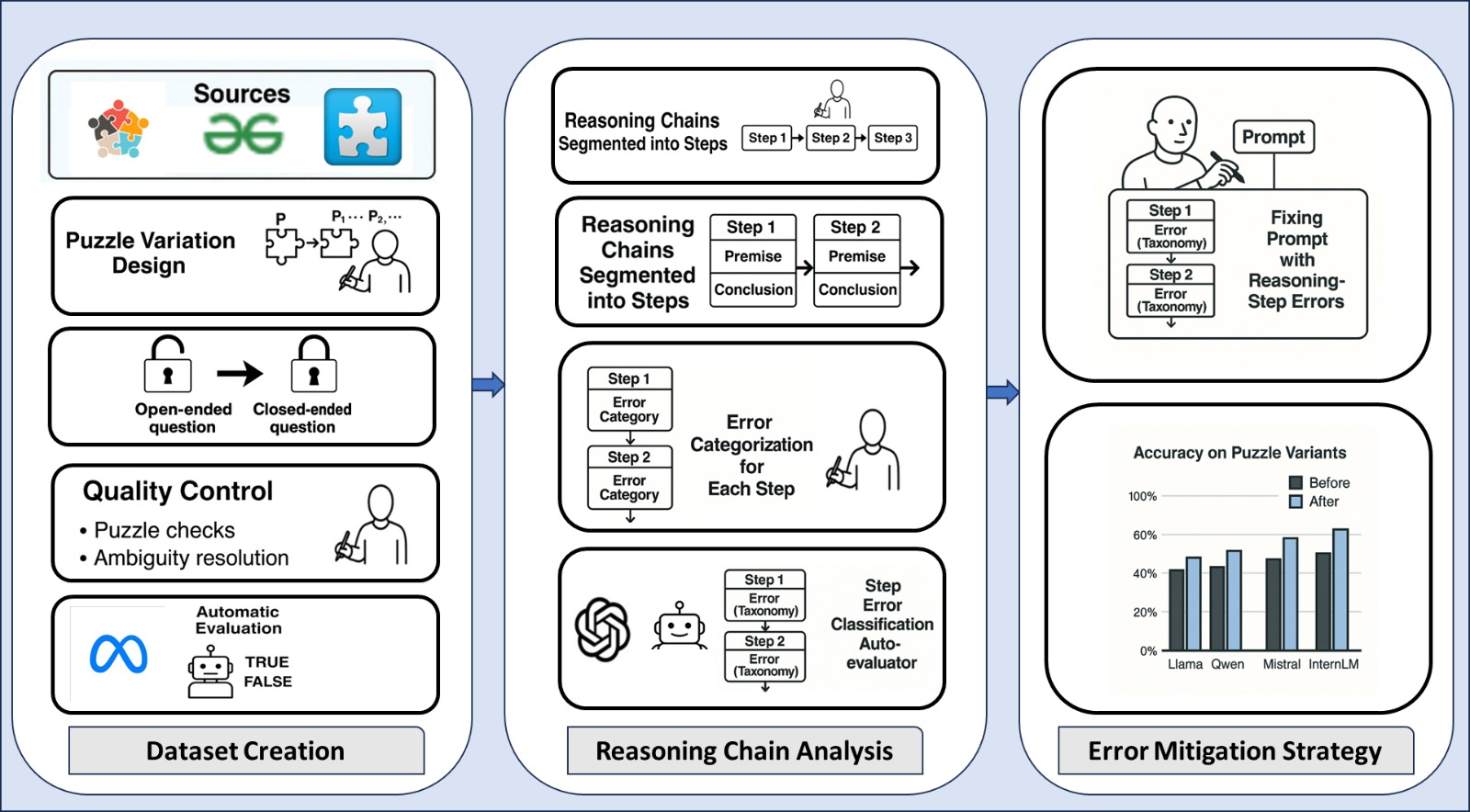}
    \caption{Overview of our three major contributions and their interconnections. 
    The benchmark enables systematic reasoning chain analysis, which informs the 
    development of error mitigation strategies. These strategies are validated 
    on the benchmark, creating a comprehensive framework.}
    \label{fig:contributions}
\end{figure*}

Using this Auto-evaluator, we analyzed a total of 50 reasoning chains generated by our five open-source models for 10 unique riddles. To validate the fidelity of this automated approach, we compared its output against our human-annotated ground truth. We randomly sampled 50 chains from our manually evaluated set and found that the Auto-evaluator achieved an average 71\% coincidence with human evaluators.  The overall architecture of our paper is shown in Fig~\ref{fig:contributions}.

\section{Experimental Setup}

\subsection{Models}
We evaluate a range of open-source LLMs, including \textbf{LLaMa} \cite{Touvron2023LLaMAOA}, \textbf{InternLM} \cite{Cai2024InternLM2TR}, \textbf{Qwen} \cite{Bai2023QwenTR}, \textbf{Phi} \cite{Abdin2024Phi3TR}, and \textbf{Mistral} \cite{Jiang2023Mistral7}, on our dataset, which consists of 25 base puzzles and a corresponding set of \textbf{perturbed variants}. Each variant is manually created by altering constraints or numerical values to change the gold answer, while preserving the underlying reasoning structure. This design forces models to recompute solutions rather than rely on memorizing answers to known problems.

To evaluate each model, we use the \textbf{Zero-shot Chain-of-Thought} setting \cite{kojima2023largelanguagemodelszeroshot}. The model receives a natural language instruction and a puzzle variant as input and is prompted to generate a reasoning chain before predicting a final answer. The evaluation was conducted on the OpenAI, Google, and Anthropic model versions released between October 2024 and May 2025, with a temperature setting of 0 for deterministic predictions. Inference for the open-source models was run locally using 2 A100 GPU units.

\subsection{Metrics}

We use \textbf{accuracy} to demonstrate the capability of LLMs in solving riddles based on their ability to predict the correct answer. To calculate this metric, we compare the LLM-generated final answers with the ground truth solution for each riddle.

The predicted answers and the ground truth solutions are typically in the form of a single word or a short phrase. We perform a normalized \textbf{Exact Match (EM)} to compare the two. Before comparison, both the predicted and the ground truth answers are standardized by converting them to lowercase and removing any leading/trailing whitespace or punctuation. An LLM's response is marked as correct only when the normalized text matches the normalized solution exactly. This process ensures that minor formatting differences do not impact the evaluation of the model's core reasoning ability.

\section{Results and Analysis}
\subsection{Objective Evaluation}

\begin{figure}[h]
    \centering
    \includegraphics[width=0.95\linewidth]{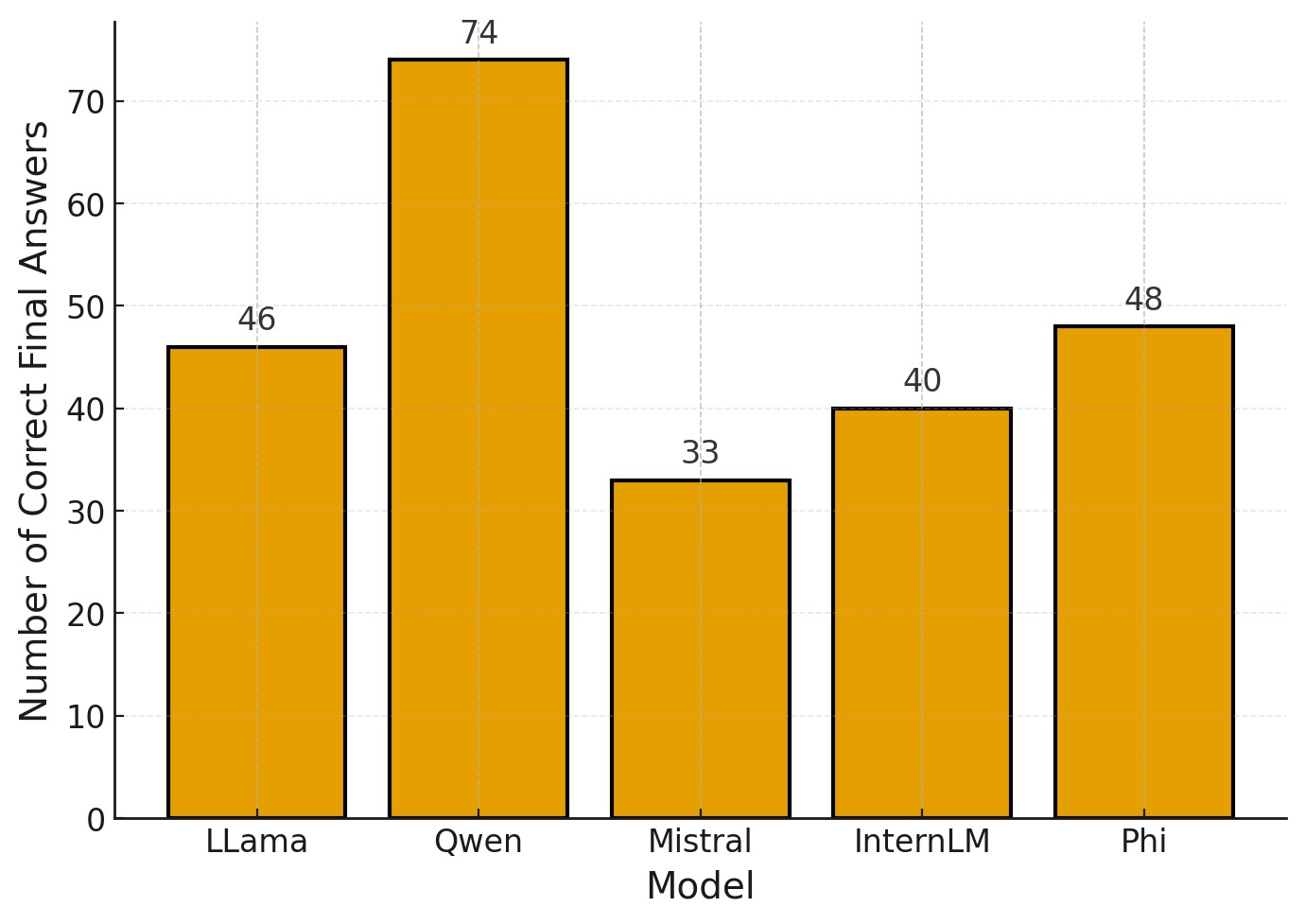}
    \caption{Performance of five different open source LLMs (LLama 3.1, Phi 4, Mistral 7B, Qwen 2.5 7B, and InternLM) in terms of accuracy on the Phantom Recall dataset.}
    \label{fig:open_source_accuracies}
\end{figure}

For the objective evaluation, our primary metric was the accuracy of the final answer provided by each model with the help of Answer verification auto-evaluator. This quantitative assessment serves as a baseline for performance, measuring a model's ability to arrive at the correct solution, irrespective of the reasoning path taken. The results, summarized in Figure \ref{fig:open_source_accuracies}, reveal a distinct performance hierarchy among the evaluated models. Qwen emerged as the top performer, achieving \textbf{74} correct final answers. It was followed by Phi and LLaMA, which secured \textbf{48} and \textbf{46} correct answers, respectively. Mistral and InternLM recorded the lowest scores in this evaluation, with Mistral providing \textbf{33} correct answers while InternLM provided \textbf{40}. These empirical results offer a clear, quantitative comparison of the models' problem-solving ability on our dataset.

\begin{figure}
    \centering
    \includegraphics[width=1\linewidth]{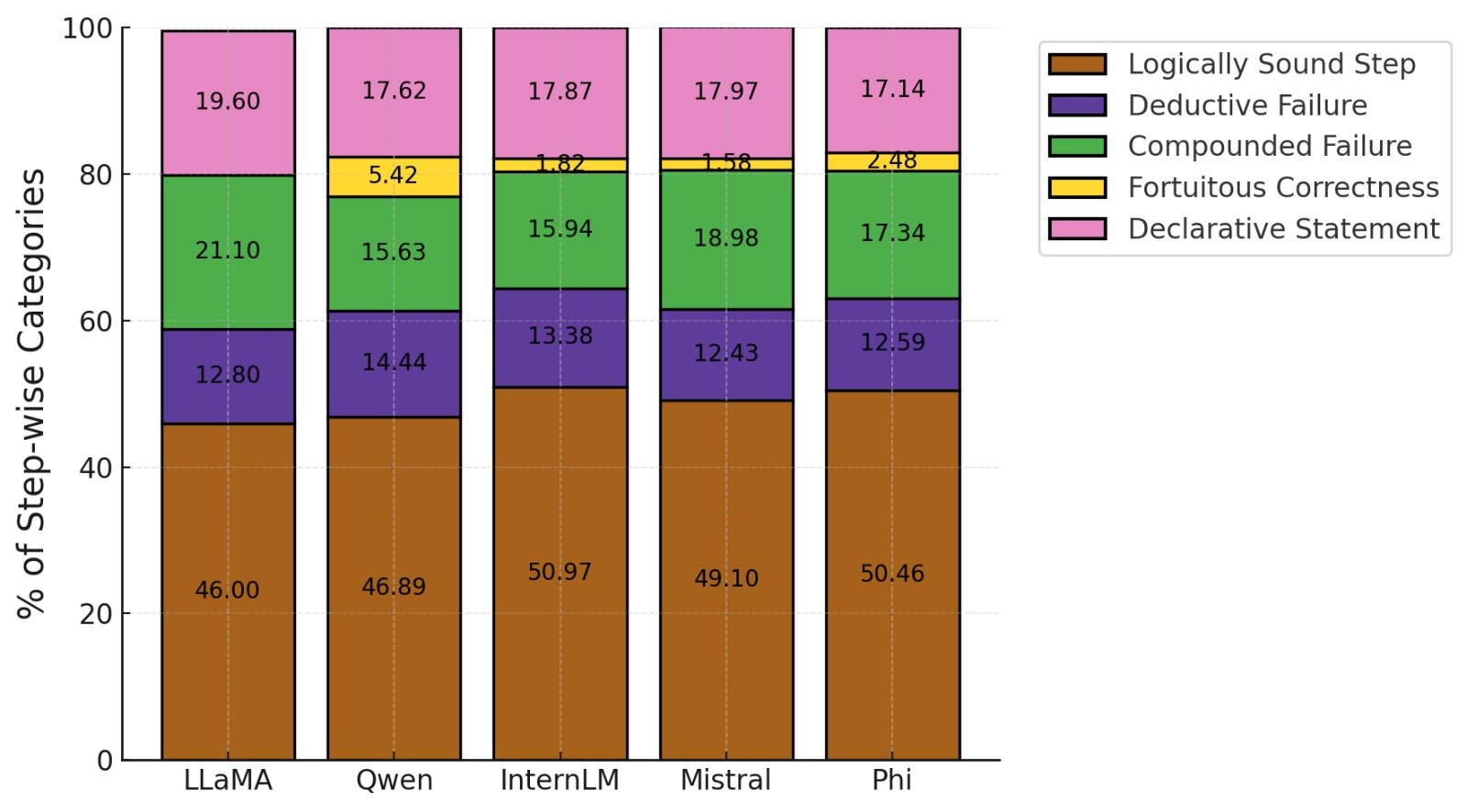}
    \caption{Comparative distribution of step-wise reasoning errors in 5 open source LLMs}
    \label{fig:stepwise-category-distribution}
\end{figure}

Figure \ref{fig:stepwise-category-distribution} compares the distribution of step-wise reasoning outcomes for five open-source LLMs--LLaMA, Qwen, InternLM, Mistral, and Phi. Each stacked bar shows the proportion of steps labeled as \emph{Logically Sound Step}, \emph{Deductive Failure}, \emph{Compounded Failure}, \emph{Fortuitous Correctness}, and \emph{Declarative Statement}. Across models, logically sound steps dominate (\(\sim\)46--51\%), indicating a substantial fraction of coherent intermediate reasoning. Deductive and compounded failures together comprise roughly 25--35\%, revealing persistent weaknesses that can derail full-chain correctness. Fortuitous correctness remains rare (<6\%), while declarative (non-reasoning) statements are relatively stable at \(\sim\)17--19\%. Overall, the figure highlights that although most steps are logically valid, deduction gaps and error propagation still limit end-to-end reasoning reliability.

\begin{figure}
    \centering
    \includegraphics[width=1\linewidth]{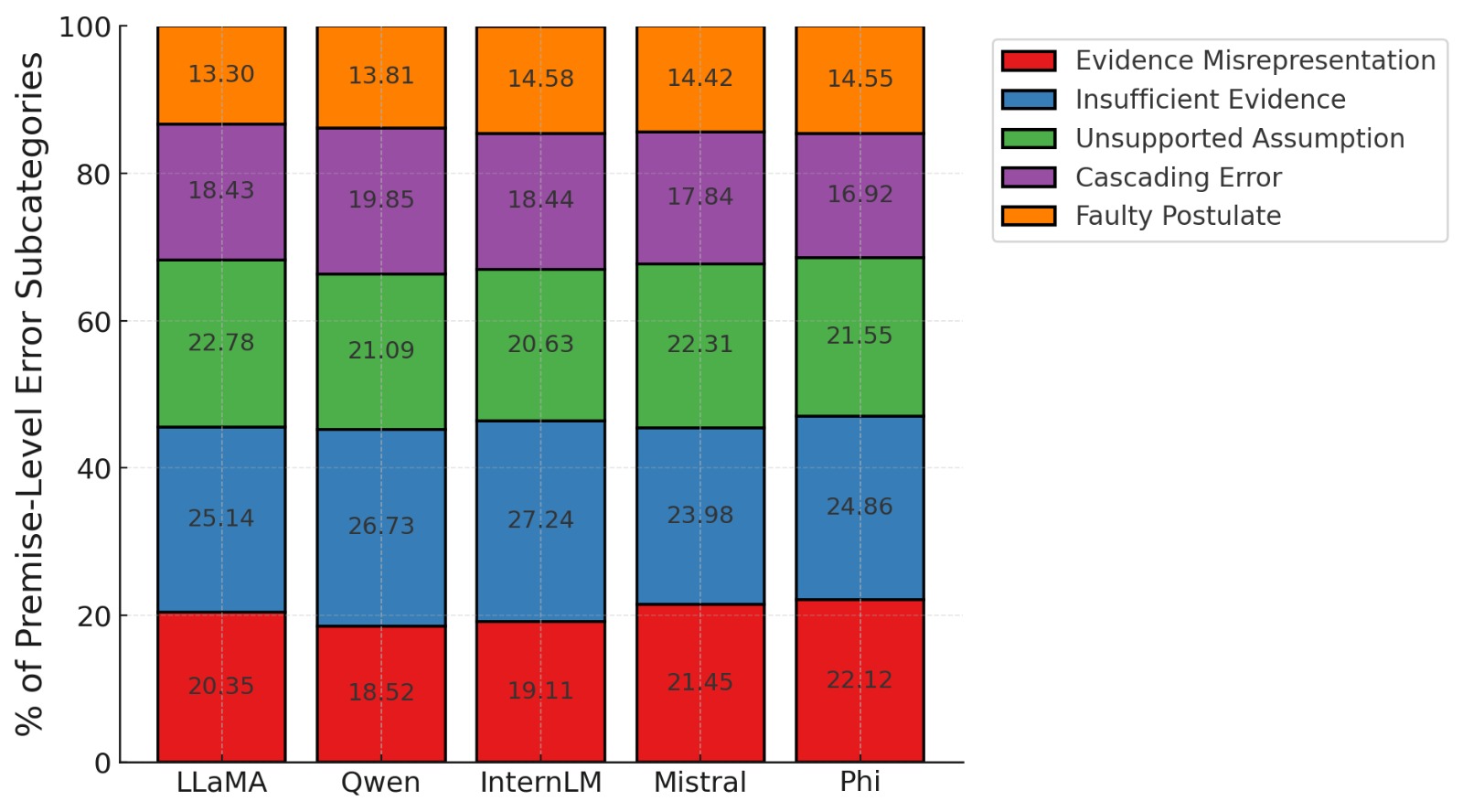}
    \caption{Comparative distribution of premise level errors in 5 open source LLMs}
    \label{fig:premise-error-distribution}
\end{figure}

\begin{figure}
    \centering
    \includegraphics[width=1\linewidth]{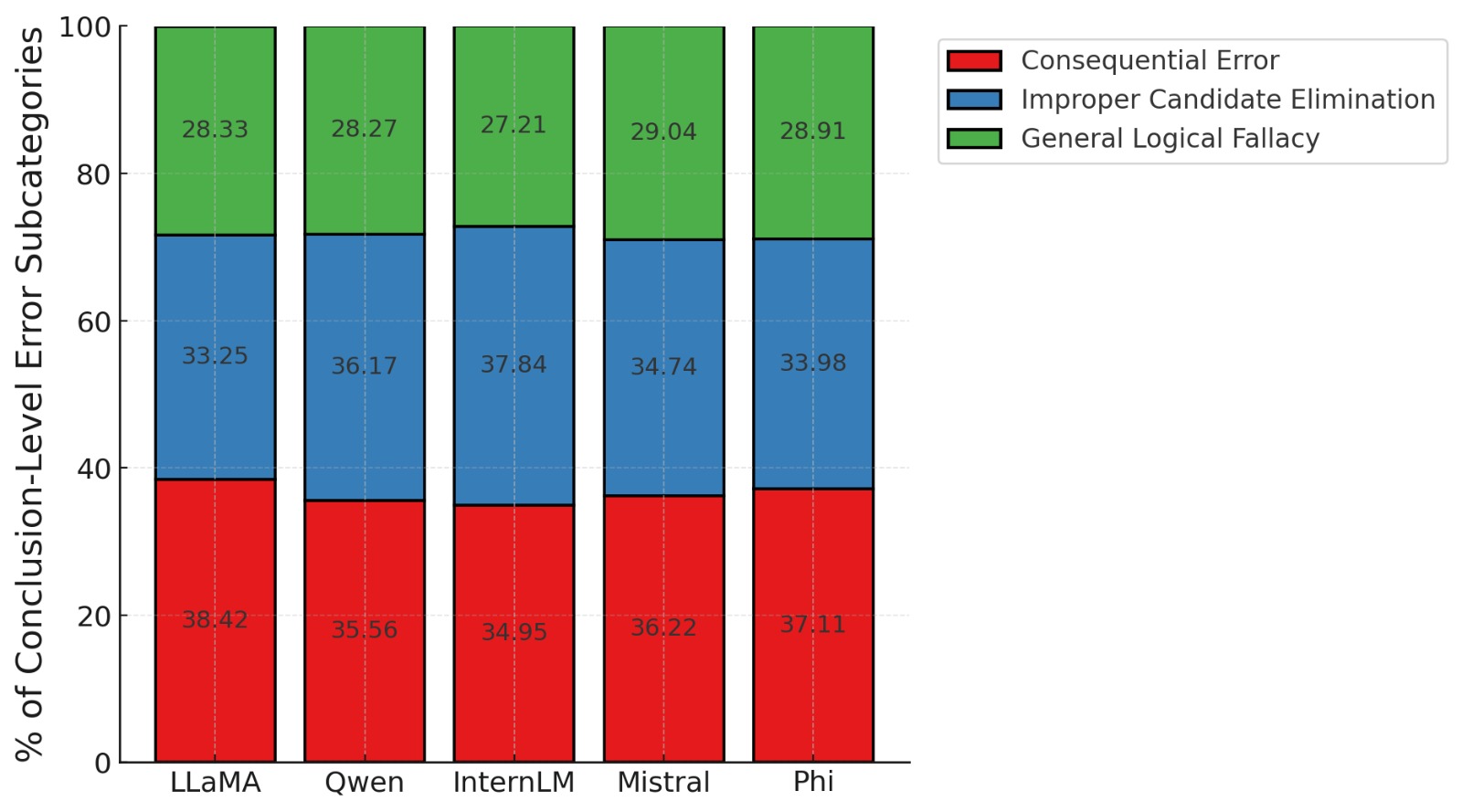}
    \caption{Comparative distribution of conclusion level errors in 5 open source LLMs}
    \label{fig:conclusion-error-distribution}
\end{figure}

Figure \ref{fig:premise-error-distribution} and Figure \ref{fig:conclusion-error-distribution} jointly reveal a consistent error pattern across LLaMA, Qwen, InternLM, Mistral, and Phi. At the \emph{premise} level (Fig.~\ref{fig:premise-error-distribution}), \emph{Insufficient Evidence} dominates (24–27\%), followed by \emph{Unsupported Assumption} (21–23\%) and \emph{Evidence Misrepresentation} (18–22\%); \emph{Cascading Error} contributes 17–20\%, while \emph{Faulty Postulate} is least common (13–15\%). At the \emph{conclusion} level (Fig.~\ref{fig:conclusion-error-distribution}), \emph{Consequential Error} is most prevalent (5–38\%), with \emph{Improper Candidate Elimination} close behind (33–38\%) and \emph{General Logical Fallacy} comprising the remainder (27–29\%). Together, the figures indicate that failures often originate from evidence insufficiency and assumption-based reasoning early in the chain, which then propagate to the final decision; among the models, InternLM (and to a lesser extent Qwen) shows slightly more balanced/robust distributions, underscoring the need for better evidence aggregation and end-of-chain verification.

\subsection{Thinking Chain Evaluation}
To systematically evaluate the logical integrity of each thinking chain, we employed a granular, step-by-step analysis. Each chain was first decomposed into its constituent steps. Following a structured annotation guideline, we designated each step as having a \textbf{'premise'} and a \textbf{'conclusion'}. This dichotomy enabled a two-level error analysis. First, the premises were scrutinized for any logical fallacies, factual inaccuracies, or reasoning flaws, which were classified as \textbf{'premise-level errors'}. Subsequently, the conclusion was evaluated, not only for its correctness but also for its logical dependency on the preceding premises. Errors identified at this final stage were classified as \textbf{'conclusion-level errors'}. This methodical approach allowed for the precise localization of failure points within the model's reasoning process.

\subsection{Discussion on Error-Mitigation Strategies}

Our primary strategy uses \textbf{prompts with explicit prohibitory instructions} to establish clear negative constraints that prune the response space. We tailored these to target specific error categories: (1) \textbf{Hallucination:} "Do not use any external knowledge" ensures faithfulness to context; (2) \textbf{Speculation:} "Do not make assumptions about missing information" prevents unwarranted inferences; (3) \textbf{Formatting:} "Do not output any commentary" and "Your response must be only the valid JSON object" enforce parsability. This shifts the model from generative to controlled behavior.

\begin{figure}[h]
    \centering
    \includegraphics[width=0.95\linewidth]{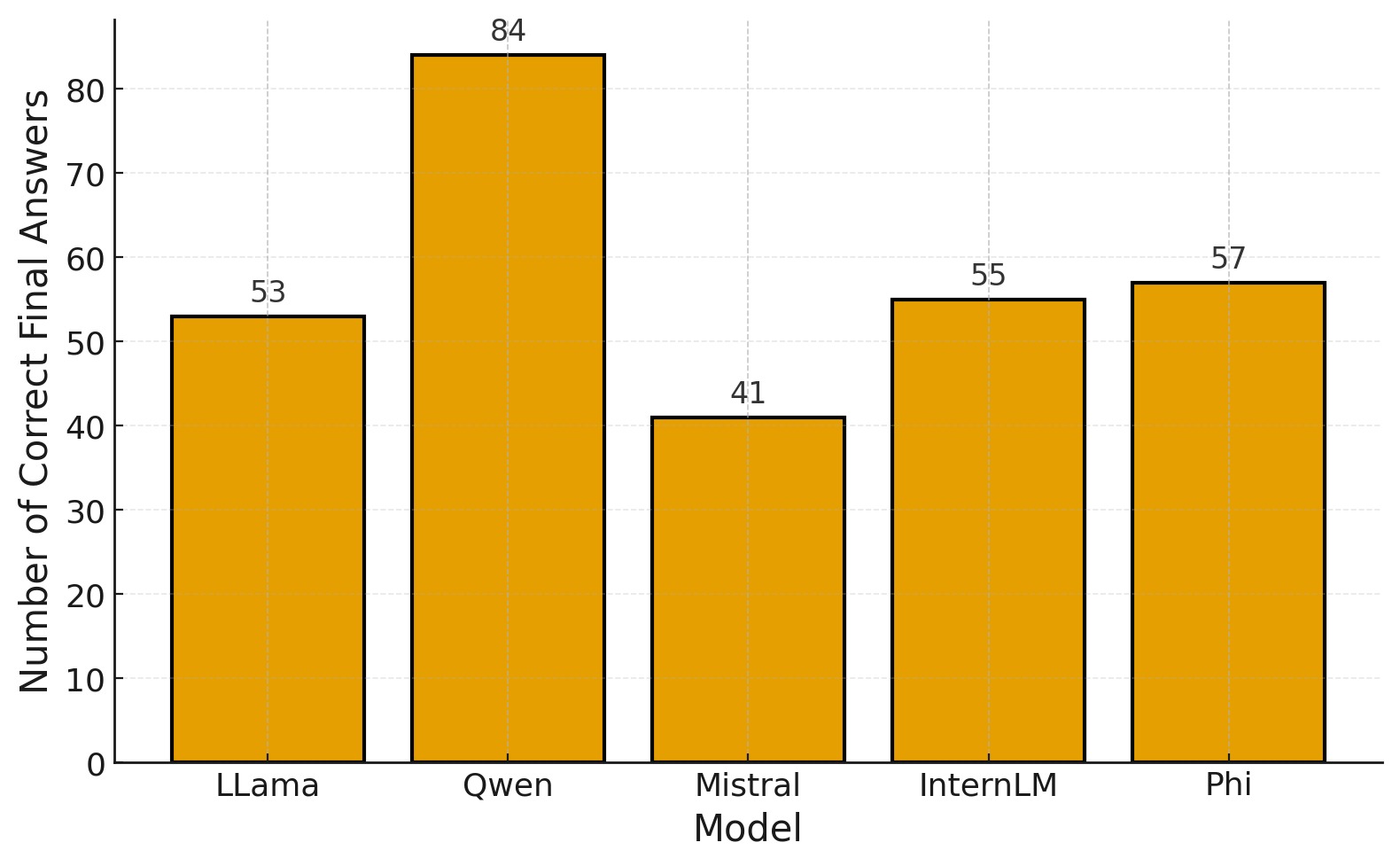}
    \caption{Performance of five different open source LLMs in terms
of accuracy on the Phantom Recall dataset with modified prompts.}
    \label{uu}
\end{figure}

Our error-mitigation strategy consistently improved accuracy across all models on the 149-puzzle dataset. Accuracy gains ranged from 4.7 percentage points (LLaMA: 30.9\%→35.6\%) to 10.1 points (InternLM: 26.8\%→36.9\%), with Qwen (49.7\%→56.4\%), Mistral (22.1\%→27.5\%), and Phi (32.2\%→38.3\%) showing intermediate improvements of 5.4–6.7 points.

\section{Conclusion}
We set out to investigate whether large language models (LLMs) have truly addressed their reasoning fragility—beyond memorized templates and benchmark familiarity. Our findings across eleven leading models suggest that while newer LLMs demonstrate marginal improvements in reasoning coherence, \textbf{they have not fundamentally solved the problem}. 
First, regarding \emph{whether LLMs have addressed these issues}, our results show that the failure mode we term \textbf{phantom recall} persists across all systems, indicating that reasoning robustness remains largely unaddressed. Second, on \emph{the extent of improvement}, performance gains are narrow and model-specific, often limited to the original puzzle formulations. Third, in examining \emph{whether models made specific fixes}, we find evidence of overfitting: some models reproduce polished but instance-bound reasoning patterns rather than general strategies. Fourth, concerning \emph{generalization}, when confronted with minimally perturbed puzzles that preserve logical structure, models rarely adjust their reasoning—demonstrating that their “fixes” lack true transferability. Finally, for \emph{perturbations to other puzzles}, the degradation in accuracy and logical soundness remains substantial even under small semantic or structural shifts.
Overall, our study reveals that while LLMs can \textbf{remember reasoning}, they often fail to \textbf{re-reason}. Their responses exhibit linguistic fluency without genuine logical flexibility. We hope that \textbf{PHANTOM RECALL} serves as both a diagnostic tool and a call to action for designing future models and training strategies that move beyond pattern matching toward context-grounded, adaptive reasoning.

\section*{Limitations}

While our benchmark provides valuable insights into LLM reasoning capabilities through logic puzzles, several limitations should be acknowledged. First, our dataset comprises 25 base puzzles with 149 systematically generated variants. Although this scale enables controlled analysis of model behavior across puzzle variations, a larger and more diverse puzzle set would strengthen claims about generalization. The current scope, however, allows for thorough manual verification and detailed error analysis that would be expensive at larger scales.

Second, we evaluate eight models, both open-source and proprietary systems. While these models represent the most widely deployed systems, the rapid pace of model development means that newer architectures may exhibit different reasoning patterns. Our selection prioritizes models with substantial adoption in research, ensuring our findings remain relevant to the broader community. 

Third, our focus on logic puzzles, while offering clear ground truth and interpretability, represents only one facet of reasoning ability. Performance on these structured tasks may not fully capture model behavior on open-ended reasoning or real-world problem-solving. However, the controlled nature of logic puzzles enables precise identification of failure modes and reasoning patterns that are difficult to isolate in less structured domains.

Finally, we acknowledge that our evaluation is conducted at a single point in time. As models continue to improve and prompting strategies evolve, the absolute performance numbers reported here may be superseded. Nevertheless, the relative comparisons between models and the qualitative insights into reasoning failures provide enduring value for understanding current limitations and informing future development.

\section*{Ethics Statement}
We have used ChatGPT and Grammarly to reword/refine statements and fix grammar mistakes.

\section*{Acknowledgements}
We thank the anonymous reviewers for their constructive suggestions. We extend our gratitude
to the Research Computing (RC), and Enterprise Technology at ASU for providing computing resources, and access to the ChatGPT enterprise version for experiments. 

\bibliography{latex/PhantomRecall}
\clearpage
\appendix
\section{Open Ended Question to Closed Ended Question}
\label{app:question_transformation}
The process of transforming riddles from open-ended to closed-ended formats was conducted through systematic iterative refinement by the two principal annotators. This transformation represents a critical methodological challenge, as it requires maintaining the essential cognitive and reasoning demands of the original riddle while fundamentally restructuring the response format to eliminate interpretive flexibility. The workflow proceeded through several carefully designed stages to ensure both consistency and validity in the transformation process.\\
Initially, one annotator proposed a modification to the final question posed in the riddle, reformulating it to constrain the response space while preserving the core reasoning challenge. This reformulation process required careful consideration of multiple factors: the question needed to eliminate the possibility of multiple valid interpretations, guide respondents toward a specific answer structure (such as binary choices, likelihood comparisons, or entity identification), and maintain the logical dependencies and inferential steps required in the original riddle. The annotator would explore various reformulation strategies, including converting "how" or "why" questions into "what is more likely" comparisons, transforming explanation requests into identification tasks, or restructuring open queries into forced-choice formats that preserved the underlying reasoning path.\\
Each modified riddle was then evaluated empirically by prompting the target LLMs and analyzing their generated responses. This evaluation phase was essential to the refinement process, as it revealed how models interpreted and responded to the reformulated questions in practice. The annotators examined these outputs systematically to identify any residual ambiguity in interpretation or reasoning paths that could lead to multiple valid answers. This examination included analyzing whether the models provided consistent answers across multiple trials, whether they exhibited uncertainty or hedging in their responses, and whether they demonstrated understanding of the intended logical structure of the riddle. Particular attention was paid to cases where models might exploit loopholes in the question wording, apply unexpected but technically valid interpretations, or demonstrate confusion about the intended answer format.\\
A riddle variant was classified as 'sufficiently closed-ended' when it satisfied one of two carefully defined criteria. First, the LLM responses demonstrated no ambiguity in interpretation, converging on a single clearly-defined answer with consistent reasoning across multiple evaluation trials. This criterion required not only that models select the correct answer, but that they do so for the right reasons, demonstrating that the question structure successfully constrained their reasoning process. Second, any remaining ambiguity could be attributed to factors independent of the question's wording itself—such as inherent semantic ambiguity in the riddle's premise, world knowledge gaps, or fundamental uncertainties in the scenario description—rather than to structural openness in the question format. This second criterion was important for distinguishing between ambiguity introduced by the transformation process and ambiguity that existed in the original riddle material itself.\\
Throughout this iterative testing-and-refinement cycle, the annotators maintained detailed records of proposed modifications, model responses, and the rationale for accepting or rejecting particular reformulations. When a proposed modification failed to adequately constrain the response space, the annotators engaged in collaborative discussion to diagnose the source of remaining ambiguity and propose alternative reformulations. This process continued until the annotators reached consensus that the closed-ended variant adequately constrained the solution space without fundamentally altering the original riddle's logical structure or cognitive demands. The consensus requirement ensured that transformations met high standards of quality and faithfulness to the source material, preventing idiosyncratic interpretations from influencing the final dataset.\\
The transformation process also involved careful consideration of how to preserve specific reasoning types across format changes. For instance, riddles requiring lateral thinking needed to maintain their misdirection elements even when converted to multiple-choice formats, while riddles based on wordplay or semantic ambiguity needed reformulations that preserved the relevant linguistic features. \\
A snapshot of the process is shown for two examples in Figure \ref{fig:question_transformation}, illustrating both the structural changes in question format and the preservation of core riddle content across the transformation.
\begin{figure*}[h]
\centering
\begin{tcolorbox}[
    colback=boxbg,
    colframe=black,
    width=\textwidth,
    arc=0mm,
    boxrule=1pt
]
\begin{tcolorbox}[
    colback=white,
    colframe=black!60,
    title=Original Version,
    boxrule=2pt,
    arc=3mm,
    fonttitle=\bfseries\small
]
\small
It's a hot rainy day and a brown hair man is standing on one side of a perennial river, with his dog on the other. There is only one option to cross the river--- a bridge. It is heavily raining outside. The man calls his dog, who immediately crosses the river without getting a single fur wet and without using a bridge. On the other hand the man is completely wet. \hl{How did the dog do it?}
\vspace{8pt}
\noindent\rule{\textwidth}{0.4pt}
\vspace{4pt}
\noindent\textbf{Question Type:} \fcolorbox{openendedtext}{openended}{\textcolor{openendedtext}{\footnotesize\bfseries~Open-Ended~}} \hfill \textbf{Response Format:} Free explanation required
\end{tcolorbox}
\vspace{10pt}
\begin{center}
\begin{tikzpicture}
    \node[font=\Large] (arrow) {$\downarrow$};
    \node[below=2pt of arrow, font=\small\itshape, text width=0.8\textwidth, align=center] 
        {Meticulous transformation to constrain response format};
\end{tikzpicture}
\end{center}
\vspace{10pt}
\begin{tcolorbox}[
    colback=white,
    colframe=black!60,
    title=Transformed Version,
    boxrule=2pt,
    arc=3mm,
    fonttitle=\bfseries\small
]
\small
It's a hot rainy day and a brown hair man is standing on one side of a perennial river, with his dog on the other. There is only one option to cross the river --- a bridge. It is heavily raining outside. The man calls his dog, who immediately crosses the river without getting a single fur wet and without using a bridge. On the other hand the man is completely wet. \hl{What is more likely: that the river is frozen or that the dog is bald?}
\vspace{8pt}
\noindent\rule{\textwidth}{0.4pt}
\vspace{4pt}
\noindent\textbf{Question Type:} \fcolorbox{closedendedtext}{closedended}{\textcolor{closedendedtext}{\footnotesize\bfseries~Closed-Ended~}} \hfill \textbf{Response Format:} Binary choice (likelihood comparison)
\end{tcolorbox}
\end{tcolorbox}

\vspace{15pt}

\begin{tcolorbox}[
    colback=boxbg,
    colframe=black,
    width=\textwidth,
    arc=0mm,
    boxrule=1pt
]
\begin{tcolorbox}[
    colback=white,
    colframe=black!60,
    title=Original Version,
    boxrule=2pt,
    arc=3mm,
    fonttitle=\bfseries\small
]
\small
A man with curly hair goes out for a walk during a rainstorm with his wife. He doesn't have any protection of his own (umbrella, hood, cap, etc.), but his wife has an umbrella big enough for them both, which they make use of. By the end of his walk, there isn't a single wet hair on his head. \hl{Why doesn't the man have wet hair?}
\vspace{8pt}
\noindent\rule{\textwidth}{0.4pt}
\vspace{4pt}
\noindent\textbf{Question Type:} \fcolorbox{openendedtext}{openended}{\textcolor{openendedtext}{\footnotesize\bfseries~Open-Ended~}} \hfill \textbf{Response Format:} Free explanation required
\end{tcolorbox}
\vspace{10pt}
\begin{center}
\begin{tikzpicture}
    \node[font=\Large] (arrow) {$\downarrow$};
    \node[below=2pt of arrow, font=\small\itshape, text width=0.8\textwidth, align=center] 
        {Meticulous transformation to constrain response format};
\end{tikzpicture}
\end{center}
\vspace{10pt}
\begin{tcolorbox}[
    colback=white,
    colframe=black!60,
    title=Transformed Version,
    boxrule=2pt,
    arc=3mm,
    fonttitle=\bfseries\small
]
\small
A man with curly hair goes out for a walk during a rainstorm with his wife. He doesn't have any protection of his own (umbrella, hood, cap, etc.), but his wife has an umbrella big enough for them both, which they make use of. By the end of his walk, there isn't a single wet hair on his head. \hl{What kept the man's hair dry?}
\vspace{8pt}
\noindent\rule{\textwidth}{0.4pt}
\vspace{4pt}
\noindent\textbf{Question Type:} \fcolorbox{closedendedtext}{closedended}{\textcolor{closedendedtext}{\footnotesize\bfseries~Closed-Ended~}} \hfill \textbf{Response Format:} Specific object/entity identification
\end{tcolorbox}
\end{tcolorbox}

\caption{Transformation of open-ended riddles into closed-ended question formats. The scenarios remain identical while the question structures shift from requiring creative explanations to constraining responses through binary choices or specific entity identification.}
\label{fig:question_transformation}
\end{figure*}
\clearpage

\section{Metrics Results from Human Evaluations}
\label{app:human_eval_metrics}

As summarized in Fig~\ref{fig:tt}, after three human evaluators independently annotated all 149 puzzles and we consolidated their judgments, o3 leads with 59.73\% accuracy (89/149), followed by Gemini~2.5~Pro at 58.39\% (87/149), Claude~Sonnet~4 at 54.36\% (81/149), Gemini~2.5~Flash at 48.32\% (72/149), Claude~3.5~Sonnet at 46.31\% (69/149), and GPT\mbox{-}4o at 29.53\% (44/149).

\begin{figure}[H]
    \centering
    \includegraphics[width=1\linewidth]{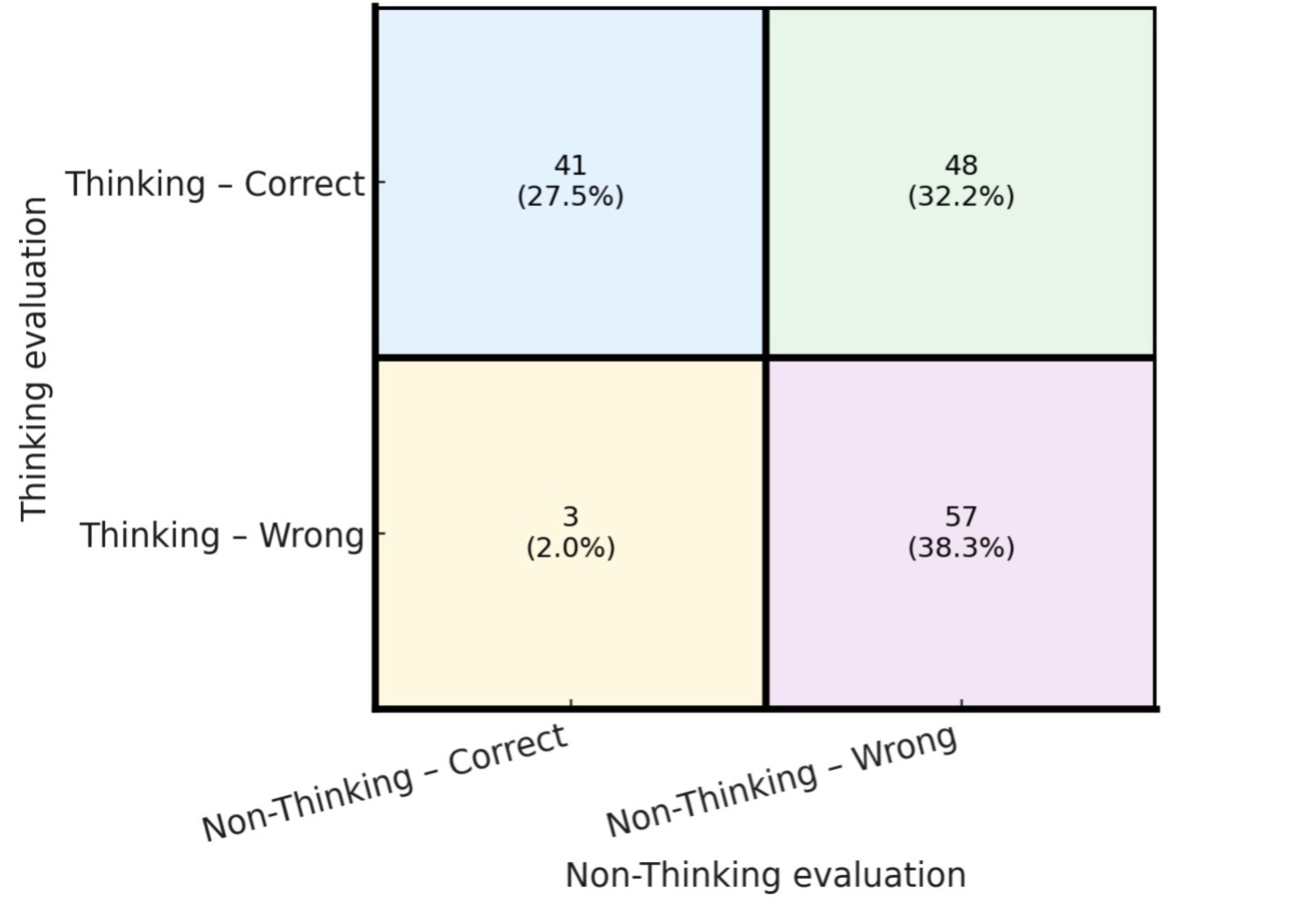}
    \caption{CHATGPT confusion matrix (n=149). Rows: Thinking outcome (Correct/Wrong); Columns: Non-Thinking outcome (Correct/Wrong). Counts and percentages are w.r.t. all puzzles. Agreement = 98/149 (65.8\%): both correct 41 (27.5\%), both wrong 57 (38.3\%); disagreements favor Thinking—48 (32.2\%) vs 3 (2.0\%).}
    \label{x1}
\end{figure}

\paragraph{Findings for \textsc{ChatGPT} (Figure~\ref{x1}).}
On the 149-puzzle set, \emph{Thinking} attains 89/149 = 59.73\% accuracy versus 44/149 = 29.53\% for \emph{Non-Thinking} (\(+30.2\) pp). Agreements total 98/149 (both correct: 41; both wrong: 57). Disagreements strongly favor Thinking—48 Thinking-only vs 3 Non-Thinking-only—indicating substantial unique coverage from chain-of-thought.

\begin{figure}[H]
    \centering
    \includegraphics[width=1\linewidth]{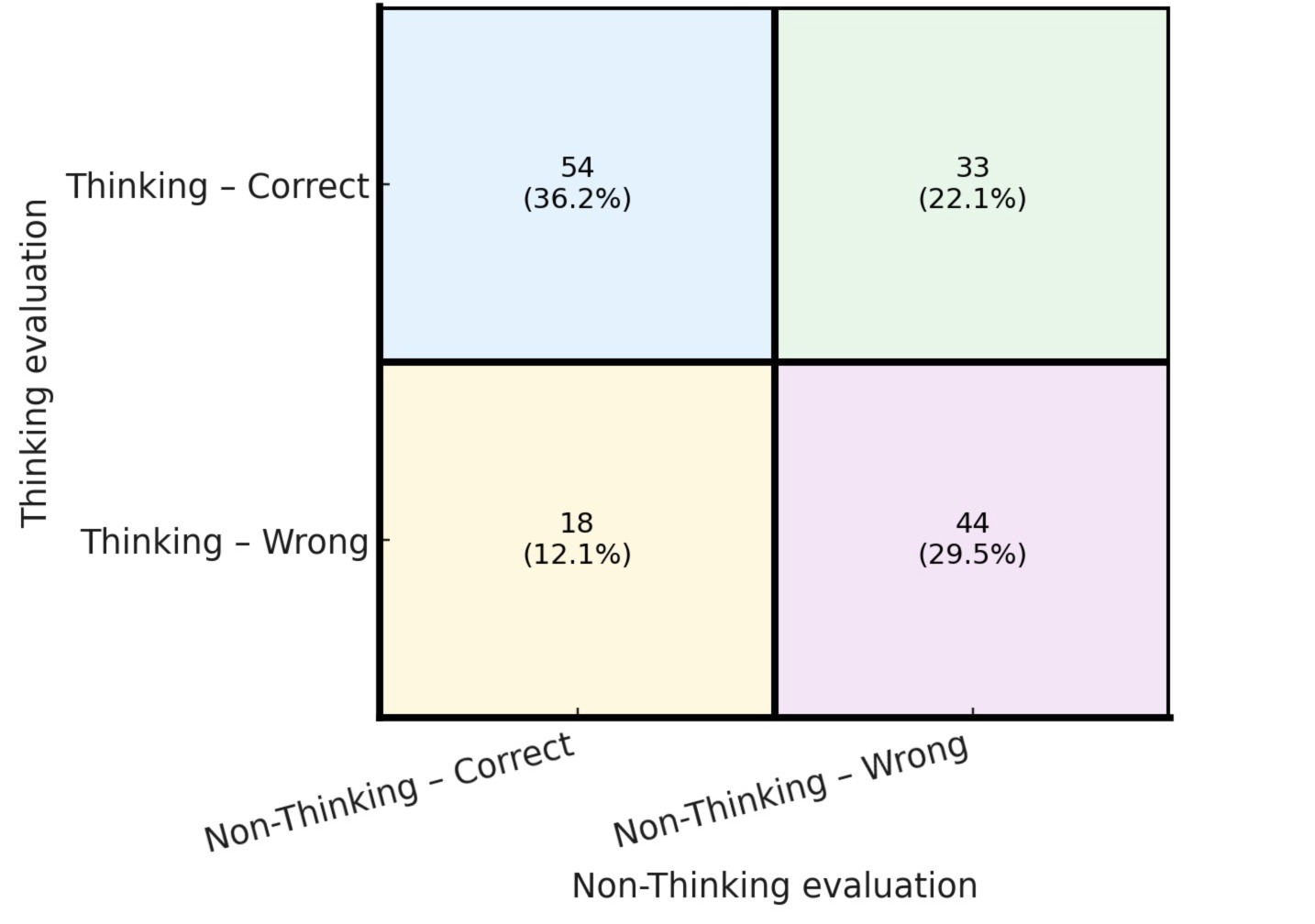}
    \caption{GEMINI confusion matrix (n=149). Rows: Thinking outcome (Correct/Wrong); Columns: Non-Thinking outcome (Correct/Wrong). Agreement = 98/149 (65.8\%): both correct 54 (36.2\%), both wrong 44 (29.5\%); disagreements: Thinking-only 33 (22.1\%) vs Non-Thinking-only 18 (12.1\%).}
    \label{x2}
\end{figure}

\begin{figure}[H]
    \centering
    \includegraphics[width=1\linewidth]{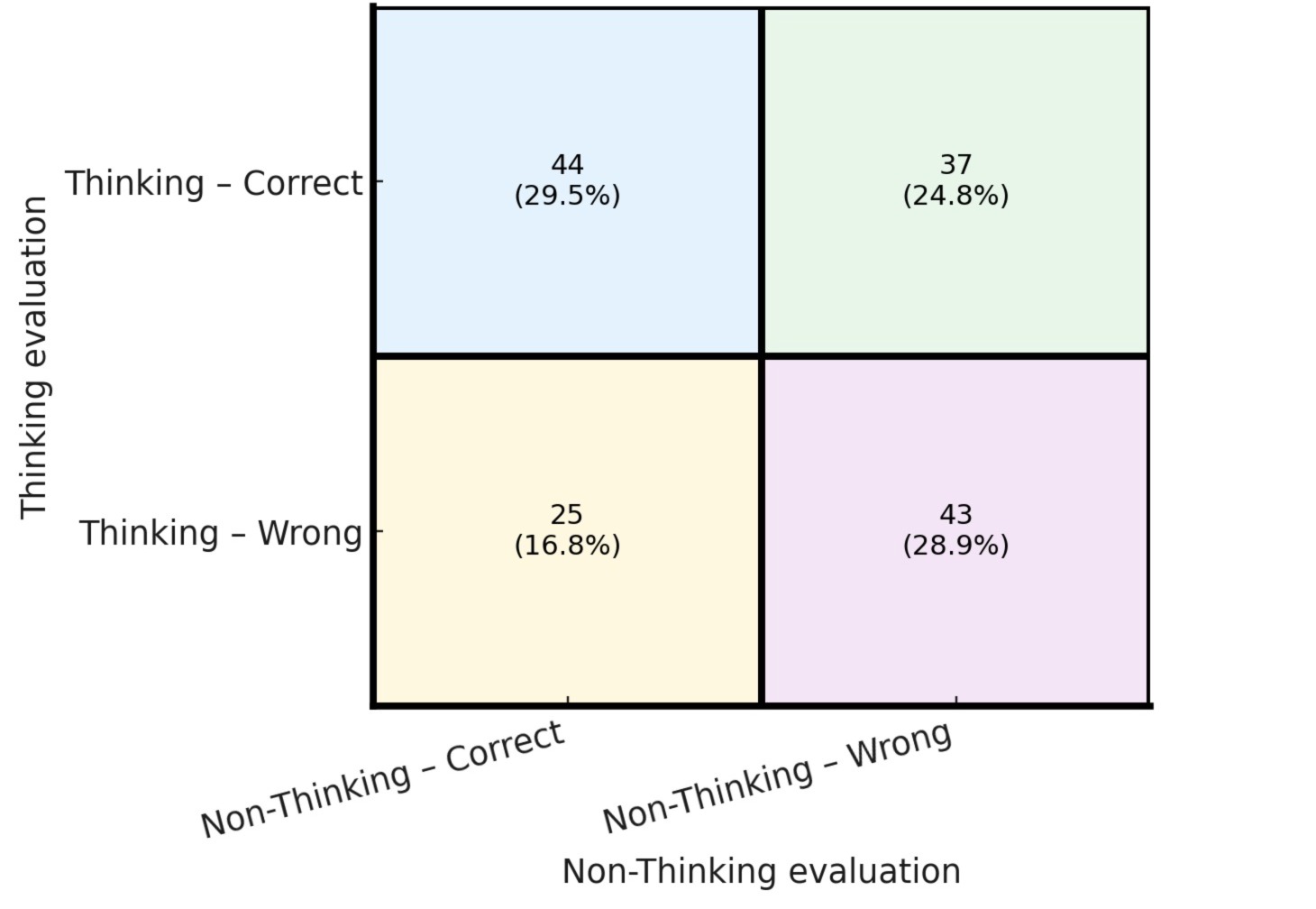}
    \caption{CLAUDE confusion matrix (n=149). Rows: Thinking outcome (Correct/Wrong); Columns: Non-Thinking outcome (Correct/Wrong). Agreement = 87/149 (58.4\%): both correct 44 (29.5\%), both wrong 43 (28.9\%); disagreements: Thinking-only 37 (24.8\%) vs Non-Thinking-only 25 (16.8\%).}
    \label{x3}
\end{figure}

\paragraph{Findings for \textsc{Gemini} (Figure~\ref{x2}).}
Thinking achieves 87/149 = 58.39\% compared to 72/149 = 48.32\% for Non-Thinking (\(+10.1\) pp). Agreements are 98/149 (both correct: 54; both wrong: 44). Disagreements show 33 Thinking-only vs 18 Non-Thinking-only (net \(+15\)), indicating a consistent but smaller lift than for ChatGPT.
\paragraph{Findings for \textsc{Claude} (Figure~\ref{x3}).}
Thinking reaches 81/149 = 54.36\% versus 69/149 = 46.31\% for Non-Thinking (\(+8.0\) pp). Agreements are lower at 87/149 (both correct: 44; both wrong: 43). Disagreements are 37 Thinking-only vs 25 Non-Thinking-only (net \(+12\)), suggesting the greatest complementarity between modes among the three models.

\clearpage
\section{Answer verification Auto-Evaluator}

\label{app:AAA}

\begin{figure}[!h]
\centering
\scalebox{0.75}{
\begin{tikzpicture}[
    node distance=0.3cm,
    every node/.style={align=center},
    box/.style={rectangle, rounded corners, draw, minimum width=2.5cm, minimum height=0.8cm, font=\small\bfseries},
    badge/.style={circle, draw, minimum size=0.7cm, font=\footnotesize\bfseries, text=white},
    rule/.style={rectangle, rounded corners=3pt, draw, minimum width=8.5cm, align=left, font=\footnotesize, inner sep=8pt}
]

\node[rectangle, rounded corners=5pt, fill=primaryblue, minimum width=10cm, minimum height=1.2cm, text=white] (header) {
    \Large\bfseries Answer Evaluation Rubric\\[2pt]
    \small Expert evaluator for puzzle question answer equivalence
};

\node[box, fill=cyan!10, draw=cyan!60, below=0.8cm of header, xshift=-3cm] (q) {Question (Q)};
\node[box, fill=green!10, draw=green!60, right=0.3cm of q] (a) {Ground Truth (A)};
\node[box, fill=yellow!20, draw=orange!60, right=0.3cm of a] (b) {Candidate (B)};

\node[below=0.8cm of a, font=\large\bfseries] (guidelines) {Evaluation Guidelines};

\node[rule, fill=truegreen!5, draw=truegreen!60, below=0.4cm of guidelines] (rule1) {
    \begin{minipage}{7.5cm}
    \textcolor{truegreen}{\textbf{\cmark TRUE Condition}}\\[3pt]
    B expresses the \textbf{same core idea and meaning as A}, even if wording differs
    \end{minipage}
};

\node[rule, fill=falsered!5, draw=falsered!60, below=0.25cm of rule1] (rule2) {
    \begin{minipage}{7.5cm}
    \textcolor{falsered}{\textbf{\xmark FALSE Conditions}}\\[3pt]
    • B changes the main concept or introduces different reasoning\\
    • B omits a key part that changes the meaning\\
    • B adds misleading or contradictory information
    \end{minipage}
};

\node[rule, fill=ignoregray!5, draw=ignoregray!60, below=0.25cm of rule2] (rule3) {
    \begin{minipage}{7.5cm}
    \textcolor{ignoregray}{\textbf{$\sim$ Ignore Trivial Differences}}\\[3pt]
    Synonyms, formatting variations, or extra neutral details
    \end{minipage}
};

\node[rule, fill=partialorange!5, draw=partialorange!60, below=0.25cm of rule3] (rule4) {
    \begin{minipage}{7.5cm}
    \textcolor{partialorange}{\textbf{! Partial Correctness}}\\[3pt]
    If B is partially correct but misses an essential component → \textbf{FALSE}
    \end{minipage}
};

\node[rule, fill=strictpurple!5, draw=strictpurple!60, below=0.25cm of rule4] (rule5) {
    \begin{minipage}{7.5cm}
    \textcolor{strictpurple}{\textbf{Strict Evaluation}}\\[3pt]
    Do NOT guess; be strict. If unsure → choose \textbf{FALSE}
    \end{minipage}
};

\node[rectangle, rounded corners=5pt, fill=primarypurple, minimum width=10cm, minimum height=1cm, text=white, below=0.5cm of rule5] (output) {
    \small OUTPUT\\[2pt]
    \Large\bfseries TRUE \;\;or\;\; FALSE
};

\end{tikzpicture}
}
\caption{Evaluation rubric for assessing conceptual equivalence between ground truth and candidate answers in puzzle question tasks.}
\label{fig:eval_rubric}
\end{figure}
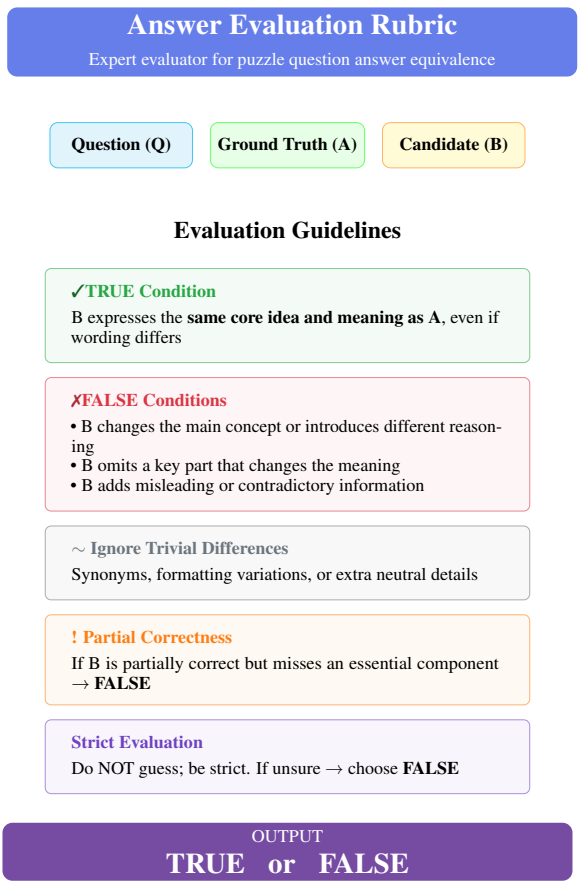

Fig~\ref{fig:eval_rubric} illustrates the rubric used by human annotators to judge whether a model’s candidate answer (\textbf{B}) is \emph{conceptually equivalent} to the ground-truth answer (\textbf{A}) for a given puzzle question (\textbf{Q}). Annotators prioritized \emph{meaning preservation} over surface form: they marked \textsc{True} only when \textbf{B} conveyed the same core idea and intended meaning as \textbf{A}, allowing for paraphrases, synonyms, formatting differences, or additional neutral details; they marked \textsc{False} whenever \textbf{B} altered the claim (e.g., different or incorrect reasoning that changes the meaning), omitted an essential component, or introduced misleading/contradictory information, with \emph{partial correctness} also treated as \textsc{False}. The instructions were deliberately strict—if uncertain, annotators selected \textsc{False} rather than guessing—yielding a binary decision (\textsc{True}/\textsc{False}) for each (\textbf{Q}, \textbf{A}, \textbf{B}) triple. Three independent human evaluators applied this rubric to all 149 items, and we consolidated their judgments into a single gold label per item, which we used for all accuracy analyses and figures in this paper.

Fig~\ref{fig:eval_examples} illustrates how we apply the conceptual-equivalence criterion using two concrete cases. Each row shows the puzzle question (Q), the gold answer (A), and the model’s candidate (B), followed by a green (\textsc{True}) or red (\textsc{False}) verdict. In \emph{Example~1}, the candidate “\emph{The dog is bald}” exactly matches the gold answer and thus preserves the intended meaning, so it is marked \textsc{True}—note that the existence of another plausible story (“the river is frozen”) is irrelevant because we judge \emph{equivalence to A}, not general plausibility. In \emph{Example~2}, the jugs puzzle already contains a jug with exactly 4L, so the gold answer is that the man can escape \emph{immediately} (i.e., zero pours). The candidate “\emph{1}” proposes a different outcome (one pour) and therefore changes the meaning; under our strict rubric (no partial credit for near misses), it is marked \textsc{False}. These examples illustrate that judgments hinge on semantic identity with the gold answer rather than surface form or alternative—but non-gold—solutions.

\begin{figure}[h] \hspace{10cm} 
\centering \resizebox{\columnwidth}{!}{ 
\begin{tikzpicture}[ node distance=0.2cm, box/.style={rectangle, rounded corners=3pt, draw, minimum width=3.2cm, minimum height=0.7cm, font=\small\bfseries, align=center}, contentbox/.style={rectangle, rounded corners=3pt, draw, text width=3.2cm, font=\footnotesize, align=left, inner sep=8pt} ] 
\node[box, fill=questioncyan!15, draw=questioncyan!80] (h1) {Question (Q)}; \node[box, fill=gtgreen!15, draw=gtgreen!80, right=0.15cm of h1] (h2) {Ground Truth (A)}; \node[box, fill=candidateyellow!20, draw=candidateyellow!80, right=0.15cm of h2] (h3) {Candidate (B)}; 
\node[contentbox, fill=questioncyan!5, draw=questioncyan!60, below=0.3cm of h1, minimum height=3.5cm] (ex1q) { A brown-haired man and his dog are on opposite sides of a river. Heavy rain. Man calls dog, who crosses without getting fur wet, no bridge. Man is wet. More likely: river frozen or dog bald? }; \node[contentbox, fill=gtgreen!5, draw=gtgreen!60, below=0.3cm of h2, minimum height=3.5cm] (ex1a) { \vspace{1cm} \textbf{The dog is bald} }; \node[contentbox, fill=candidateyellow!5, draw=candidateyellow!60, below=0.3cm of h3, minimum height=3.5cm] (ex1c) { \vspace{1cm} \textbf{The dog is bald} }; 
\node[rectangle, rounded corners=2pt, fill=gtgreen, text=white, font=\footnotesize\bfseries, below=0.1cm of ex1c] (eval1) {\cmark TRUE}; 
\draw[thick, gray!40] ([yshift=-0.35cm]ex1q.south west) -- ([yshift=-0.35cm]ex1c.south east); 
\node[contentbox, fill=questioncyan!5, draw=questioncyan!60, below=0.65cm of ex1q, minimum height=3.5cm] (ex2q) { Man locked in room with 3 jugs (8L, 5L, 4L) filled with 8L, 4L, 2L respectively. Needs exactly 4L in one jug to escape. Minimum pours? }; \node[contentbox, fill=gtgreen!5, draw=gtgreen!60, below=0.65cm of ex1a, minimum height=3.5cm] (ex2a) { \vspace{0.5cm} \textbf{The man can escape immediately} }; \node[contentbox, fill=candidateyellow!5, draw=candidateyellow!60, below=0.65cm of ex1c, minimum height=3.5cm] (ex2c) { \vspace{1.2cm} \textbf{1} }; 
\node[rectangle, rounded corners=2pt, fill=red!80, text=white, font=\footnotesize\bfseries, below=0.1cm of ex2c] (eval2) {\xmark FALSE}; \end{tikzpicture} } \caption{Examples of answer evaluation. \textbf{Example 1:} Candidate matches ground truth conceptually (TRUE). \textbf{Example 2:} Candidate answer "1" misses the key insight that one jug already contains 4L, making escape immediate without any pours (FALSE).} \label{fig:eval_examples} \end{figure}
\clearpage
\section{Average Number of Steps Generated by LLMs}
\label{app:step_counts}
For incorrect answers, QWEN produces by far the longest chains of thought—51.1 steps on average (range 5–179), indicating a tendency to “over-reason” when it’s off-track. In contrast, LLAMA is the most concise and consistent, with the lowest average of 5.0 steps and a tight 4–10 range. MISTRAL (7.8, 4–10) and INTERNLM (7.6, 5–13) show moderate length and variability, while PHI (8.8, 2–29) exhibits broader dispersion despite a modest mean. Overall, these results suggest that more steps do not equate to better outcomes—when models are wrong, longer reasoning often reflects digressions rather than recovery, whereas shorter, steadier traces (e.g., LLAMA) indicate more bounded but still unsuccessful reasoning. The information discussed is presented, for readability, in \textbf{Table \ref{tab:llm-steps}}.
\begin{table}[!h]
\centering
\caption{Average number of reasoning steps generated by different LLMs for incorrect answers.}
\label{tab:llm-steps}
\setlength{\tabcolsep}{5pt}
\renewcommand{\arraystretch}{1.05}
\resizebox{\columnwidth}{!}{%
\begin{tabular}{@{}lrrr@{}}
\toprule
\textbf{Model} & \textbf{Average Steps} & \textbf{Min Steps} & \textbf{Max Steps} \\
\midrule
MISTRAL  & 7.8  & 4  & 10  \\
PHI      & 8.8  & 2  & 29  \\
QWEN     & 51.1 & 5  & 179 \\
LLAMA    & 5.0  & 4  & 10  \\
INTERNLM & 7.6  & 5  & 13  \\
\bottomrule
\end{tabular}%
}
\end{table}

\section{Human Annotation Guidelines and Process}
\label{app:annotation_guidelines}
This appendix provides a comprehensive description of our human annotation methodology for identifying and categorizing errors in open source LLM-generated reasoning chains. The annotation process was carefully designed to ensure high-quality, reliable labels that serve as ground truth for evaluating our proposed error detection framework. We recruited three independent human evaluators for this task, each bringing complementary expertise to ensure comprehensive and nuanced error detection. The following structured guidelines were provided to all human evaluators and served as the primary reference throughout the annotation process:
\begin{tcolorbox}[
colback=blue!15!white,
colframe=blue!50!black,
title=\textbf{Step-by-Step Annotation Protocol},
fonttitle=\bfseries,
coltitle=white,
rounded corners,
boxrule=1.5pt,
arc=4mm
]
\small
\begin{enumerate}[noitemsep]
    \item Take a reasoning chain that has been broken down into steps
    
    \item Break each step into premise and conclusion
    
    \item Classify the step into one of the \textbf{5 Main Categories}:
    \begin{itemize}[noitemsep]
        \item \textbf{Logically Sound Step:}Correct conclusion validly derived from correct premise
        \item \textbf{Deductive Failure:}Correct premise leads to incorrect conclusion due to flawed reasoning
        \item \textbf{Compounded Failure:}Incorrect premise leads to incorrect conclusion
        \item \textbf{Fortuitous Correctness:}Flawed premise coincidentally results in correct conclusion
        \item \textbf{Declarative Statement:}Restates a clue or previous finding without drawing new conclusion
    \end{itemize}
    
    \item If errors exist, identify the \textbf{Premise level Error} (if any):
    \begin{itemize}[noitemsep]
        \item Evidence Misrepresentation
        \item Insufficient Evidence
        \item Unsupported Assumption
        \item Cascading Error
        \item Faulty Postulate
    \end{itemize}
    
    \item If errors exist, identify the \textbf{Conclusion level Error} (if any):
    \begin{itemize}
        \item Consequential Error
        \item Improper Candidate Elimination
        \item General Logical Fallacy
    \end{itemize}
\end{enumerate}
\vspace{5pt}
\textbf{Important Notes:}
\begin{itemize}[leftmargin=*]
    \item A single reasoning step may contain multiple errors at the same level or across both levels
    
    \item When uncertain, err on the side of marking an error if the reasoning appears questionable or insufficiently justified
    
    \item \textit{Cascading Error} should be marked only when the current step's premise directly inherits an error from a previous step's faulty conclusion
    
    \item Provide brief justification for each annotation decision in the notes field
    
    \item Consider both explicit and implicit assumptions when evaluating premises
\end{itemize}
\end{tcolorbox}
The pairwise coincidence (percent agreement) among the three human evaluators is \(A_{12}=90\%\), \(A_{13}=93\%\), and \(A_{23}=89\%\). The average pairwise coincidence is therefore \(\bar{A}=\tfrac{A_{12}+A_{13}+A_{23}}{3}=\tfrac{90+93+89}{3}=90.67\%\) (approximately \(90.7\%\)), indicating high consistency across annotators (range \(89\%\)--\(93\%\)).
\clearpage
\section{The LLM-based Step Error Classification Autoevalutor}
\label{app:auto_evaluator}

This appendix describes the architecture and implementation of our LLM-based auto-evaluator for reasoning step analysis.

\subsection{Architecture Overview}
\label{app:evaluator_architecture}

\begin{figure}[!h]
\centering
\begin{minipage}{0.45\textwidth}
\begin{tikzpicture}[
    scale=0.45,
    node distance=0.3cm and 0.5cm,
    box/.style={rectangle, draw=blue!60, line width=0.6pt, minimum width=1.2cm, minimum height=0.5cm, align=center, font=\tiny, fill=blue!15, rounded corners=1pt},
    input/.style={rectangle, draw=green!60, line width=0.6pt, minimum width=1cm, minimum height=0.45cm, align=center, fill=green!20, font=\tiny, rounded corners=1pt},
    llm/.style={rectangle, draw=orange!70, line width=0.7pt, minimum width=1.3cm, minimum height=0.55cm, align=center, fill=orange!25, font=\tiny\bfseries, rounded corners=1pt},
    output/.style={rectangle, draw=purple!60, line width=0.6pt, minimum width=1cm, minimum height=0.45cm, align=center, fill=purple!20, font=\tiny, rounded corners=1pt},
    arrow/.style={-{Stealth[length=1.2mm]}, line width=0.5pt, draw=black!70}
]
\fill[blue!3, rounded corners=2pt] (-2,-2) rectangle (14,2);
\node[input] (reasoning) {Reasoning Chain};
\node[llm, right=0.4cm of reasoning] (phase1) {Phase 1:\\Deconstruct};
\node[box, above right=0.15cm and 0.6cm of phase1] (premise) {Premise $P$};
\node[box, below right=0.15cm and 0.6cm of phase1] (conclusion) {Conclusion $C$};
\node[llm, right=1.5cm of phase1] (phase2) {Phase 2:\\Validate};
\node[output, right=0.4cm of phase2] (output) {Output\\Results};
\draw[arrow] (reasoning) -- (phase1);
\draw[arrow] (phase1.15) -- ++(0.2,0) |- (premise.west);
\draw[arrow] (phase1.-15) -- ++(0.2,0) |- (conclusion.west);
\draw[arrow] (premise.east) to[out=0, in=90] (phase2.north);
\draw[arrow] (conclusion.east) to[out=0, in=270] (phase2.south);
\draw[arrow, draw=orange!70] (phase2.west) .. controls +(-1.8,0.8) and +(-1.8,-0.8) .. (phase2.west);
\draw[arrow] (phase2) -- (output);
\end{tikzpicture}
\caption{Two-phase reasoning validation pipeline. 
Phase 1 extracts premise and conclusion 
from the reasoning chain, which 
Phase 2 then validates with 
iterative refinement (loop).}
\label{fig:pipeline}
\end{minipage}
\end{figure}
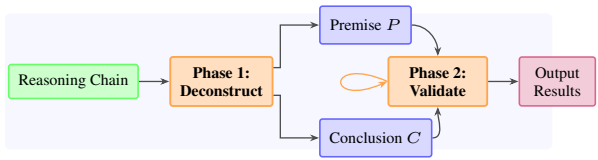

Our auto-evaluator employs a two-agent pipeline to systematically analyze each reasoning step (Figure~\ref{fig:pipeline}). Each agent performs a specialized function:

\begin{enumerate}
    \item \textbf{Deconstructor:} Parses a reasoning step into its foundational premise and resulting conclusion.
    \item \textbf{Validator:} Evaluates the correctness of both the premise (against context) and the conclusion (given the premise).
    \item \textbf{Classifier:} Maps the validation results to our error taxonomy.
\end{enumerate}

\subsection{Agent Prompts}
\label{app:evaluator_prompts}
The Agent Prompts section provides the complete prompt specifications for each component of our two-phase auto-evaluator system. As described earlier, the auto-evaluator employs a two-agent pipeline to systematically analyze reasoning chain steps, with each agent performing a specialized function in the evaluation process. The prompts detailed in this section define the specific instructions, input formats, and expected outputs for the Deconstructor, Validator, and Classifier agents, ensuring that the evaluation process is both rigorous and replicable.\\
The Deconstructor prompt (Figure~\ref{fig:agent1-prompt}) guides the agent to parse each reasoning step into its foundational premise and resulting conclusion, establishing the atomic units for subsequent evaluation. This prompt instructs the model to identify the key information being assumed or referenced (the premise) and the logical inference or claim being made (the conclusion), providing structured output that facilitates downstream validation. The prompt explicitly defines the distinction between premises and conclusions, providing examples of each to ensure consistent interpretation across diverse reasoning steps. Additionally, the Deconstructor is instructed to preserve the semantic content of the original step while reformulating it into the premise-conclusion structure, maintaining fidelity to the model's original reasoning process.\\
The Validator prompt (Figure~\ref{fig:agent2-prompt}) specifies how to assess the correctness of both the extracted premise (against the provided context) and the conclusion (given the premise), producing structured validation results. The prompt defines clear criteria for evaluating premise validity—whether the assumed information is supported by the given context—and conclusion soundness—whether the inference logically follows from the premise. The validator produces binary judgments for each component along with justifications for its assessments. Critically, the prompt emphasizes that premise validation should be performed strictly against the provided context without incorporating external knowledge, while conclusion validation should assess logical entailment independent of whether the premise itself is correct. This separation allows us to distinguish between errors stemming from faulty assumptions versus errors in logical inference. The prompt also includes detailed instructions for handling edge cases, such as implicit premises, multi-part conclusions, and reasoning steps that draw on multiple contextual elements.\\
Finally, the Classifier prompt (Figure~\ref{fig:agent3-prompt}) maps these validation outcomes to our error taxonomy, enabling systematic categorization of reasoning failures. Based on the premise and conclusion validity from the Validator, the Classifier assigns each erroneous step to one of our predefined error categories. The prompt provides the complete error taxonomy with definitions and examples, ensuring classification decisions are grounded in our theoretical framework. Together, these prompts form the operational backbone of our automated evaluation framework, ensuring consistent and reproducible analysis across all reasoning chains in our dataset.
\begin{figure*}[t]
  \centering
  \footnotesize
  \setstretch{0.85}
  
  \noindent
  \begin{tcolorbox}[title=Agent 1: Deconstructor, enhanced, colback=blue!5, 
                     colframe=blue!60!black, coltitle=white, fonttitle=\bfseries,
                     width=\textwidth, fontupper=\footnotesize, left=2mm, right=2mm,
                     breakable]
\small
You are a precise logical analysis tool. Your task is to deconstruct a 
reasoning step into its foundational premise and its resulting conclusion.

\textbf{Context:}

\texttt{\{context\}}

\textbf{Reasoning Step:}

\texttt{\{reasoning\_step\}}

Based on the reasoning step, identify the premise (the starting fact or 
assumption) and the conclusion (the inference drawn from it). Provide your 
answer in JSON format only, with the keys "premise" and "conclusion". Do not 
include any explanatory text or markdown formatting.
  \end{tcolorbox}

\caption{Agent 1 (Deconstructor): Deconstructs reasoning steps into premise and conclusion components.}
\label{fig:agent1-prompt}
\end{figure*}

\begin{figure*}[h]
  \centering
  \footnotesize
  \setstretch{0.85}
  
  \noindent
  \begin{tcolorbox}[title=Agent 2: Validator, enhanced, colback=green!5, 
                     colframe=green!60!black, coltitle=white, fonttitle=\bfseries,
                     width=\textwidth, fontupper=\footnotesize, left=2mm, right=2mm,
                     breakable]
\small
You are a meticulous logical validator. You will receive a context, a premise, 
and a conclusion. Evaluate each part and provide a structured JSON response.

\textbf{Error Sub-categories:}

For wrong premises: (1) Evidence Misrepresentation (2) Insufficient Evidence 
(3) Unsupported Assumption (4) Cascading Error (5) Faulty Postulate

For wrong conclusions: (a) Consequential Error (b) Improper Candidate 
Elimination (c) General Logical Fallacy

\textbf{Context:}

\texttt{\{context\}}

\textbf{Analysis Input:}

Premise: \texttt{\{premise\}}

Conclusion: \texttt{\{conclusion\}}

\textbf{Instructions:}

1. \textbf{Evaluate the Premise:} Compare the premise ONLY against the provided
   context. Is it factually correct according to the context? If it's 
   wrong, which sub-category best explains why? Determine if its status 
   is "Right Premise" or "Wrong Premise".

2. \textbf{Evaluate the Conclusion:} Assuming the premise is TRUE, does the 
   conclusion logically follow from it? If the reasoning is flawed, which 
   sub-category best explains why? Determine if its status is "Right 
   Conclusion" or "Wrong Conclusion".

Provide your analysis in the following JSON format ONLY. Your entire response 
must be a single, valid JSON object, starting with \{ and ending with \}.

\begin{verbatim}
{
  "premise_evaluation": {
    "status": "Right Premise OR Wrong Premise",
    "reason_code": "1-5 or N/A",
    "analysis": "Briefly explain your reasoning."
  },
  "conclusion_evaluation": {
    "status": "Right Conclusion OR Wrong Conclusion",
    "reason_code": "a-c or N/A",
    "analysis": "Briefly explain if the conclusion follows."
  }
}
\end{verbatim}
  \end{tcolorbox}

\caption{Agent 2 (Validator): Validates premise against context and evaluates logical consistency of conclusions.}
\label{fig:agent2-prompt}
\end{figure*}

\begin{figure*}[t]
  \centering
  \footnotesize
  \setstretch{0.85}
  
  \noindent
  \begin{tcolorbox}[title=Agent 3: Classifier, enhanced, colback=red!5, 
                     colframe=red!60!black, coltitle=white, fonttitle=\bfseries,
                     width=\textwidth, fontupper=\footnotesize, left=2mm, right=2mm,
                     breakable]
\small
You are an error classification system. Convert the provided validation report 
into a final error code based on the given taxonomy. Respond with the final 
code only.

\textbf{Taxonomy:}

Compounded Failure: Wrong Premise \& Wrong Conclusion

Fortuitous Correctness: Wrong Premise, Right Conclusion

Deductive Failure: Right Premise, Wrong Conclusion

Logically Sound Step: Right Premise, Right Conclusion

Declarative Statement: No Conclusion / pure restatement

\textbf{Validation Report:}

\texttt{\{report\_str\}}

\textbf{Instructions:}

1. Determine the broad category (Compounded Failure, Fortuitous Correctness,
   Deductive Failure, Logically Sound Step, Declarative Statement) from the
   status fields.
   
2. Append the reason codes in parentheses.

3. If the premise is right, omit its reason code.

4. If the conclusion is right, omit its reason code.

5. Format: Category(PremiseCode, ConclusionCode) or Category(Code).

Respond with only the final code.
  \end{tcolorbox}

\caption{Agent 3 (Classifier): Classifies validation results into final error taxonomy codes.}
\label{fig:agent3-prompt}
\end{figure*}
\FloatBarrier
\clearpage
\section{Effectiveness of Error Mitigation}
The effectiveness of our error mitigation strategy is demonstrated through the design and implementation of a carefully structured prompt template that embodies the principles of explicit prohibitory instructions. Our prompt engineering process was characterized by \textbf{meticulous iterative refinement and strengthening}, where each component was systematically tested, evaluated, and enhanced based on observed failure modes across diverse problem sets.
\subsection{Prompt Architecture and Design Rationale}
The final prompt template employs a multi-layered architecture that guides the model through a structured problem-solving protocol while simultaneously imposing strict constraints on its behavior. The template is organized into three primary phases:
\textbf{Phase 1: Understanding (Step 1 - UNDERSTAND).} This initial phase explicitly instructs the model to engage in careful problem comprehension by reading the problem twice, identifying the specific question being asked, and enumerating all given information. This deliberate repetition serves to anchor the model's attention on the problem statement itself rather than pattern-matching to similar problems from training data.
\textbf{Phase 2: Solution Generation (Step 2 - SOLVE).} The solving phase emphasizes transparency and incrementality, requiring the model to work step-by-step, show all calculations and logical reasoning, and perform inline verification at each step. This structured approach reduces the likelihood of logical leaps or omitted reasoning that could introduce errors.
\textbf{Phase 3: Verification (Step 3 - VERIFY).} The final phase implements a systematic checking mechanism with three specific validation criteria: confirming the answer addresses the actual question posed, verifying correct units and format, and validating the solution by substitution back into the original problem.
\subsection{Strategic Use of Prohibitory Instructions}
A distinguishing feature of our approach is the \textbf{STRICT RULES} section, which explicitly delineates prohibited behaviors using visual markers (\xmark) for constraints and approved behaviors (\cmark) for acceptable fallback strategies. The prohibitions specifically target the most common failure modes observed during development:
\begin{itemize}
\item \textbf{External knowledge restriction:} "NO external knowledge or memorized solutions" prevents the model from applying domain-specific facts not present in the problem statement, directly addressing hallucination tendencies.
\item \textbf{Assumption prevention:} "NO assumptions about unstated information" and "NO assuming 'standard' interpretations unless explicitly stated" combat the model's tendency to fill gaps with plausible but unwarranted information.
\item \textbf{Procedural rigor:} "NO skipping steps in calculations'' enforces complete working, making errors more detectable and reasoning more transparent.
\item \textbf{Explicit fallback mechanism:} The instruction to state "Cannot solve: missing [X]" when information is ambiguous provides the model with a legitimate alternative to guessing, reducing false confidence in incomplete solutions.
\end{itemize}
\subsection{Iterative Refinement Process}
The development of this prompt template involved systematic strengthening through multiple iterations. Initial versions used general instructions like "solve carefully" or "show your work", which proved insufficient for error mitigation. Through empirical testing on benchmark problems, we identified specific failure patterns—such as models assuming standard values (e.g., gravitational acceleration) without being told, or answering related but different questions than what was asked.
Each identified failure mode prompted the addition or strengthening of a corresponding constraint. For example, early versions lacked the "re-read the original question" instruction in the CRITICAL section, leading to instances where models would solve a problem correctly but report an answer in the wrong format or for a subtly different question. The addition of this explicit re-verification step, with its three-point checklist, reduced such format and question-mismatch errors substantially.
The visual distinction between prohibited (\xmark) and approved (\cmark) behaviors was introduced after observing that purely textual negative constraints were sometimes overlooked. The use of visual markers increased the salience of these instructions, improving adherence rates.
\subsection{Performance Impact}
As noted previously, this prompt methodology achieved approximately \textbf{10\% improvement on closed-source LLMs} and \textbf{just over 5\% improvement on open-source LLMs} compared to baseline prompting approaches. The performance gains were particularly pronounced in problems requiring:
\begin{itemize}[noitemsep]
\item Careful distinction between given and assumed information
\item Multi-step reasoning with intermediate verification
\item Precise adherence to output format specifications
\item Recognition of insufficient information to solve a problem
\end{itemize}
Notably, the structured three-phase protocol combined with explicit prohibitions proved more effective than either component alone, suggesting a synergistic effect between positive guidance (the step-by-step structure) and negative constraints (the prohibitory rules). The final "CRITICAL" section, which forces a terminal verification pass before finalizing the answer, served as a crucial safety net that caught errors that survived earlier checking phases.
This comprehensive approach demonstrates that effective error mitigation in LLMs requires not merely instructing models on what to do, but equally importantly, explicitly constraining what they should \textit{not} do, coupled with structured protocols that enforce verification at multiple stages of the problem-solving process.
\clearpage
\section{Some examples of puzzles where most of the closed source models fail}
\begin{figure}[h!]
    \centering
    \begin{subfigure}[t]{\textwidth}
        \centering
        \includegraphics[width=\textwidth]{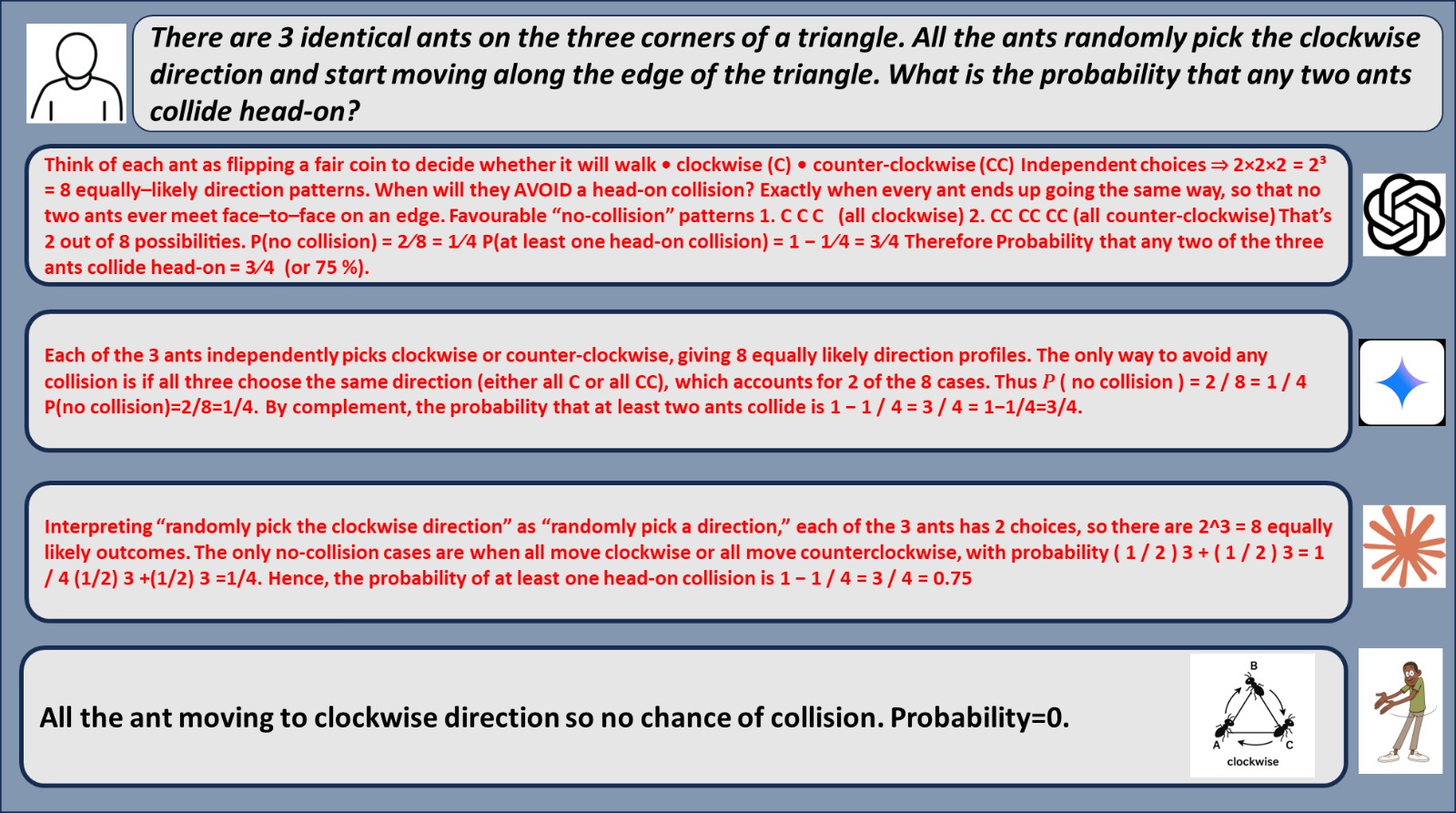}
        \caption{Puzzle variation where GPT, Gemini, and Claude fail.}
        \label{fig:var-a}
    \end{subfigure}
    \vspace{0.6em}
    \begin{subfigure}[t]{\textwidth}
        \centering
        \includegraphics[width=\textwidth]{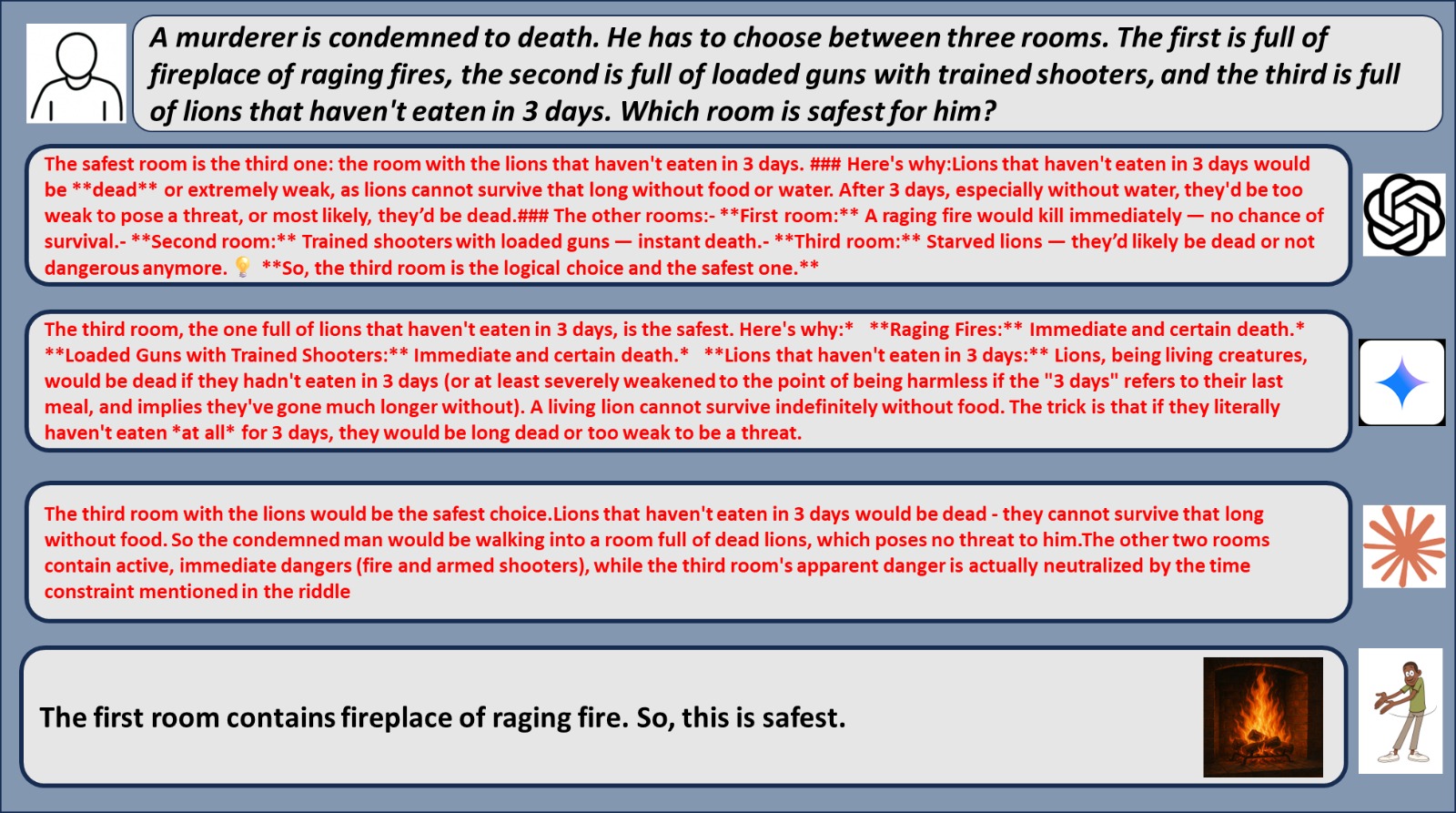}
        \caption{Puzzle variation where GPT, Gemini, and Claude fail.}
        \label{fig:var-b}
    \end{subfigure}
\end{figure}

\begin{figure*}[p]
    \ContinuedFloat
    \centering
    \begin{subfigure}[t]{\textwidth}
        \centering
        \includegraphics[width=\textwidth]{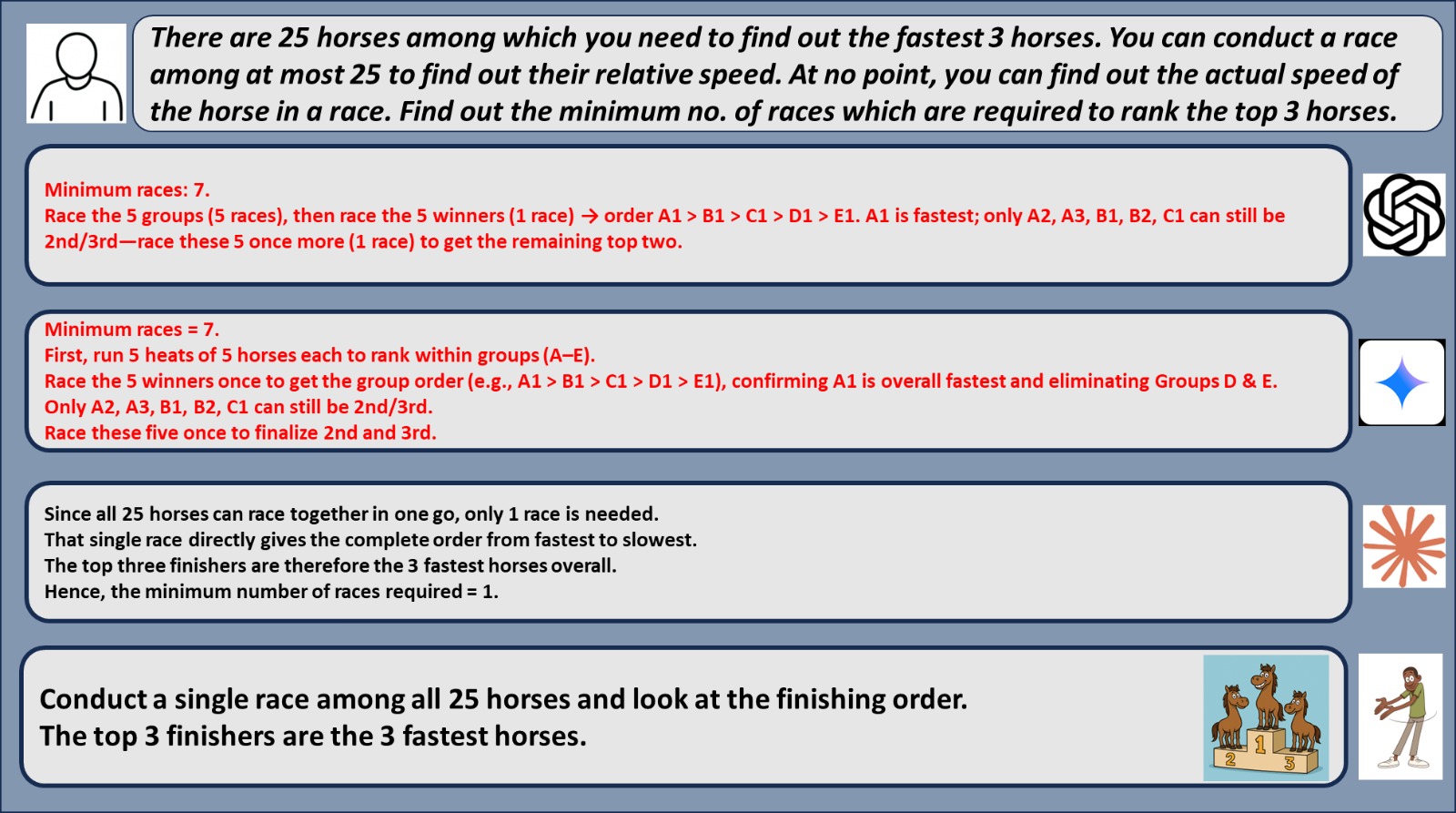}
        \caption{Variation where GPT, Gemini fail but Claude succeeds.}
        \label{fig:var-c}
    \end{subfigure}
    \vspace{0.6em}
    \begin{subfigure}[t]{\textwidth}
        \centering
        \includegraphics[width=\textwidth]{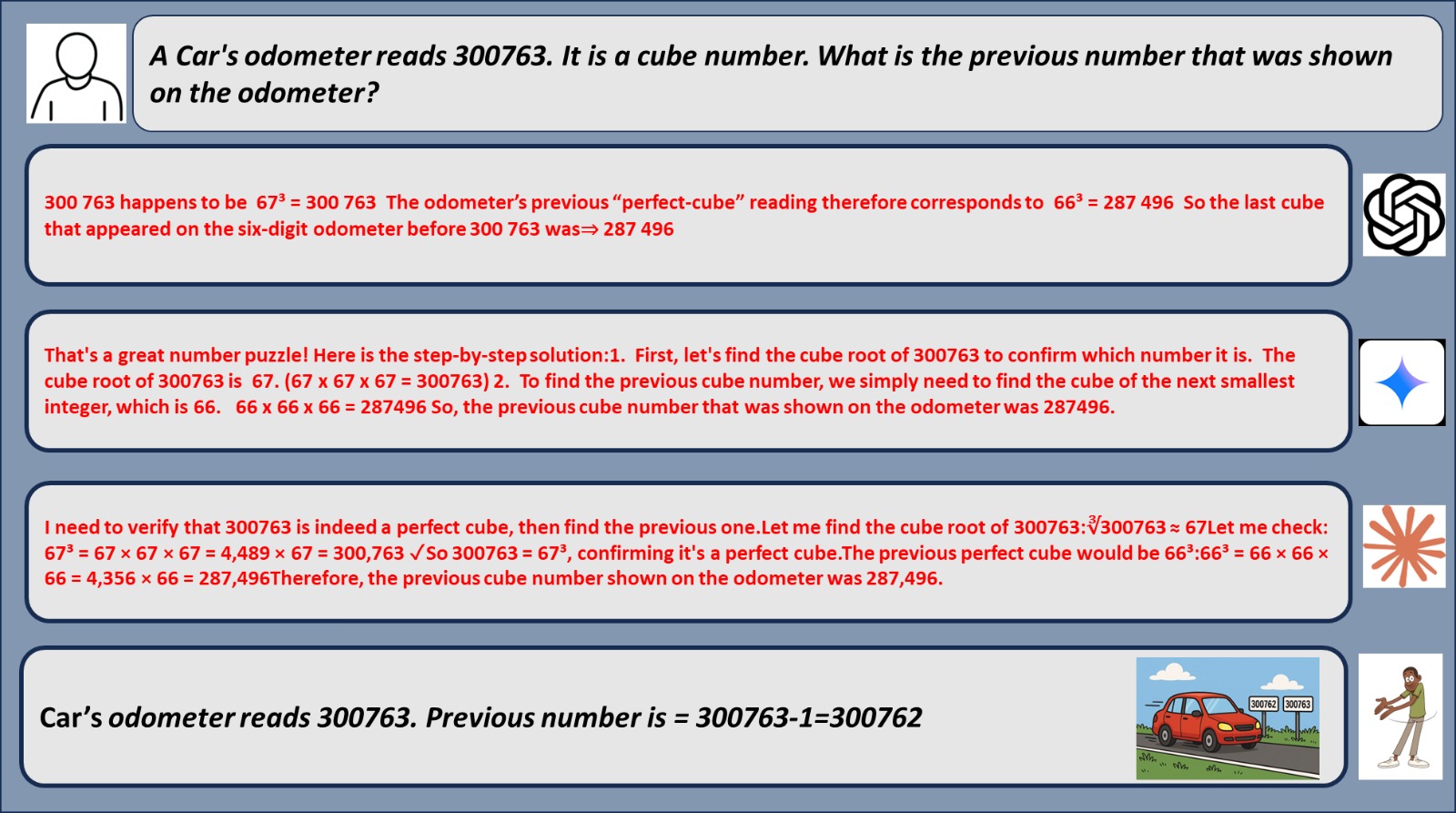}
        \caption{Variation where all three models fail.}
        \label{fig:var-d}
    \end{subfigure}
    \caption{Examples of puzzle variations from our dataset.}
    \label{fig:puzzle-variations}
\end{figure*}
\end{document}